\PassOptionsToPackage{dvipsnames}{xcolor} 

\documentclass[12pt]{wlscirep}

\usepackage[sort&compress,numbers]{natbib}

\usepackage[utf8]{inputenc}
\usepackage[T1]{fontenc}
\usepackage{textcomp}
\usepackage{courier} 

\usepackage{graphicx}    
\usepackage{multirow}    
\usepackage{amsmath,amssymb,amsfonts}  
\usepackage{amsthm}      
\usepackage{mathrsfs}    
\usepackage[title]{appendix} 
\usepackage{manyfoot}    
\usepackage{booktabs}    

\usepackage[linesnumbered,ruled,vlined]{algorithm2e}
\SetKwComment{Commentm}{// }{}  
\usepackage{algorithmicx}
\usepackage{algpseudocode}
\usepackage{float} 

\usepackage{listings}  

\usepackage{cleveref} 
\usepackage{xspace}

\usepackage[most]{tcolorbox}
\tcbuselibrary{breakable}

\usepackage{longtable}
\usepackage{booktabs}
\usepackage{placeins}
\usepackage{adjustbox}

\newcommand{\method}{KUMO\xspace}

\usepackage{longtable}
\usepackage{booktabs}

\begin{document}

\title{Generative evaluation of complex reasoning in large language models}

\author[1,$\dag$]{Haowei Lin}
\author[1,$\dag$]{Xiangyu Wang}
\author[1,$\dag$]{Ruilin Yan}
\author[2]{Baizhou Huang}
\author[5]{Haotian Ye}
\author[2]{Jianhua Zhu}
\author[1]{Zihao Wang}
\author[5]{James Zou}
\author[3,4]{Jianzhu Ma}
\author[1]{Yitao Liang}
\affil[1]{Institute for Artificial Intelligence, Peking University, Beijing, China.}
\affil[2]{Wangxuan institute of computer technology, Peking University, Beijing, China.}
\affil[3]{Department of Electronic Engineering, Tsinghua University, Beijing, China.}
\affil[4]{Institute for AI Industry Research, Tsinghua University, Beijing, China.}
\affil[5]{Computer Science Department, Stanford University, California, United States.}
\affil[$\dag$]{Equal contribution.}
\affil[*]{Correspondence should be addressed to: 
    \href{mailto:majianzhu@tsinghua.edu.cn}{majianzhu@tsinghua.edu.cn},
    \href{mailto:yitaol@pku.edu.cn}{yitaol@pku.edu.cn}.
}

\begin{abstract}
With powerful large language models (LLMs) demonstrating superhuman reasoning capabilities, a critical question arises: Do LLMs genuinely reason, or do they merely recall answers from their extensive, web-scraped training datasets? Publicly released benchmarks inevitably become contaminated once incorporated into subsequent LLM training sets, undermining their reliability as faithful assessments. To address this, we introduce \method, a generative evaluation framework designed specifically for assessing reasoning in LLMs. \method synergistically combines LLMs with symbolic engines to dynamically produce diverse, multi-turn reasoning tasks that are partially observable and adjustable in difficulty. Through an automated pipeline, \method continuously generates novel tasks across open-ended domains, compelling models to demonstrate genuine generalization rather than memorization. We evaluated 23 state-of-the-art LLMs on 5,200 tasks across 100 domains created by \method with easy and hard settings, benchmarking their reasoning abilities against university students. Our findings reveal that many LLMs have outperformed university-level performance on easy reasoning tasks, and the ``reasoning LLMs'' that think in chains-of-thought before answering reach university-level performance on complex reasoning challenges. These findings suggest that LLMs are capable of genuine reasoning, particularly as \method tasks are resistant to memorization.  Moreover, LLM performance on \method tasks correlates strongly with results on newly released real-world reasoning benchmarks, underscoring \method's value as a robust, enduring assessment tool for genuine LLM reasoning capabilities.
\end{abstract}

\maketitle

\label{sec:intro}

Reasoning—the cognitive process of using evidence, arguments, and logic to reach conclusions—is fundamental to problem-solving, decision-making, and critical thinking~\citep{Wason1972PsychologyOR,Fagin1988ReasoningAK}. Enhancing the reasoning capabilities of Artificial Intelligence (AI) has been a core goal since the field’s inception, dating back to early developments such as automatic theorem provers and expert systems built with search algorithms and graphical models~\citep{Polu2020GenerativeLM, jiang2022thor, jiang2023draft, pmlr-v97-yang19a, Lauritzen1990LocalCW, Pederson2001ProbabilisticNA, Castillo1996ExpertSA, Sheikhtaheri2014DevelopingAU, Neapolitan2012ProbabilisticRI}. In the deep learning era, AI systems have exhibited increasingly powerful reasoning abilities. For instance, AlphaZero demonstrated superhuman performance in strategic games like chess, shogi, and Go~\citep{Silver2018AGR}; AlphaGeometry outperformed the average gold medalist in solving Olympiad-level geometry problems~\citep{AlphaGeometryTrinh2024}; and OpenAI’s o3 model achieved a gold medal at the 2024 International Olympiad in Informatics (IOI)~\citep{ElKishky2025CompetitivePW}. Among these advancements, large language models (LLMs) have emerged as a particularly significant component, enabling impressive reasoning capabilities across a wide range of tasks. Looking ahead, the prospect of LLMs achieving superhuman intelligence in even more domains is highly promising.

However, there remains an ongoing debate regarding whether LLMs genuinely reason or merely recall answers derived from their extensive, web-scraped training datasets. Many researchers are involved in the debate~\citep{Ahmed2025,Gajre2023}. Addressing this debate, or further advancing the development of reasoning capabilities in LLMs, first necessitates establishing a suitable evaluation mechanism, which itself is a non-trivial challenge. Recall that reasoning, by definition, is ``the cognitive process of using evidence, arguments, and logic to reach conclusions''. Thus, the essence of reasoning lies not merely in arriving at valid conclusions, but in the reasoning process itself. However, directly evaluating this cognitive process presents significant challenges. Firstly, cognitive processes are generally invisible, and in deep neural networks, interpreting computational flows underlying an LLM’s reasoning remains difficult. Secondly, even though reasoning LLMs such as OpenAI’s o1~\citep{ElKishky2025CompetitivePW} and o3~\citep{ElKishky2025CompetitivePW}, or DeepSeek’s R1~\citep{DeepSeekAI2025DeepSeekR1IR}, can explicitly articulate their reasoning “thoughts”~\citep{DeepSeekAI2025DeepSeekR1IR, DBLP:journals/corr/abs-2410-13639, Zhong2024EvaluationOO, Qin2024O1RJ}, multiple valid reasoning paths may exist that lead to the same correct conclusion~\citep{Stanovich_West_2000, Evans2010IntuitionAR, wang2023selfconsistency}. Furthermore, parsing reasoning processes articulated in natural language is inherently complex. Recent work such as MR-Ben~\citep{Zeng2024MRBenAM} demonstrates that even state-of-the-art LLMs struggle to identify errors within reasoning chains, further underscoring the difficulty of accurately assessing reasoning processes. Consequently, we argue that, for the time being, it is more practical to use conclusions as proxies for evaluating reasoning performance in LLMs, and we term this as conclusion-based evaluation.

To date, numerous conclusion-based evaluation benchmarks have been proposed for assessing LLM reasoning. Existing benchmarks such as LogiQA~\citep{logiqa}, LogiQA 2.0~\citep{Liu2023LogiQA2I}, ReClor~\citep{Yu2020ReClor:}, AR-LSAT~\citep{Wang2021FromLT}, ConTRoL~\citep{Li2024FormalLLMIF}, and AGIEval~\citep{zhong-etal-2024-agieval} are primarily derived from standardized tests (e.g., Chinese National Civil Service Exam, GMAT, LSAT, GRE, recruitment exams). These sources provide high-quality, expert-designed logical reasoning questions at considerable scale. Other datasets, such as LINGOLY~\citep{bean2024lingoly}, FOLIO~\citep{han-etal-2024-folio}, CLUTRR~\citep{Sinha2019CLUTRRAD}, and GSM8K~\citep{Cobbe2021TrainingVT}, are either crafted by domain experts or constructed through crowd-sourcing efforts. With explicit correct conclusion annotation, evaluating reasoning performance is straightforward for conclusion-based benchmarks by comparing model-generated answers to the annotated ground truths.
However, it is important to emphasize certain risks associated with conclusion-based evaluation. Since conclusions serve merely as proxies for actual reasoning performance, the fidelity of this evaluation approach is contingent upon the conclusion being genuinely derived through reasoning processes rather than memorization. Memorization can easily occur due to \emph{dataset contamination}~\citep{dong-etal-2024-generalization, Roberts2023DataCT, balloccu-etal-2024-leak, aiyappa-etal-2023-trust}. Once datasets become publicly available, their content is susceptible to being incorporated into the pre-training datasets of LLMs.
Recent evidence of such contamination includes observations that LLM performance on Codeforces problems sharply declines for problems published after an LLM's training cutoff date~\citep{jain2025livecodebench, Roberts2024ToTC}. Conversely, performance prior to the cutoff date strongly correlates with the frequency of a problem's appearance on GitHub~\citep{Roberts2024ToTC}. Additionally, a recently developed hand-crafted variant of the widely used math dataset GSM8K has revealed that several models have likely overfit to this benchmark~\citep{Cobbe2021TrainingVT, zhang2024a}. To counteract contamination, one recent benchmark, LiveBench~\citep{white2025livebench}, attempts to avoid contamination by updating questions monthly, but this demands substantial human effort. Considering these challenges, there is a clear need for a benchmark that can be efficiently updated, effectively addressing dataset contamination while balancing resource demands.

In this paper, we introduce the concept of \emph{generative evaluation} as a dynamic benchmarking method, in which benchmark questions are algorithmically generated rather than manually curated. Prior studies, including LogicBench~\citep{jain2025livecodebench} and DYVAL~\citep{Zhu2023DyValDE}, have applied generative methods primarily to evaluate logical expression solving and query processing tasks on synthetic directed acyclic graphs (DAGs). Such synthetic data can be efficiently compiled and synthesized, naturally supporting continuously updatable benchmarks. Inspired by these rule-based generative approaches, we propose an advanced generative evaluation framework called \method, designed specifically for assessing complex reasoning capabilities in LLMs. \method involves a structured reasoning game wherein a participant interacts iteratively with a system to gather evidence and draw conclusions within partially observable environments. The scenarios in this game are contextually rich and emulate real-world reasoning tasks across various domains, such as medical diagnostics, educational assessment, and chemical material detection. To ensure the evaluation isolates reasoning performance from domain-specific knowledge, each task includes a knowledge guidebook provided to the participants. Tasks within \method are dynamically generated through an automated neural-symbolic pipeline, where advanced LLMs collaborate closely with an SAT-based symbolic engine. This integration allows precise control over problem complexity across multiple dimensions, resulting in a benchmarking approach that is both highly versatile and difficult to saturate.

In our experiments, we randomly generate 5,200 tasks across 100 domains with two difficulty levels using \method to evaluate the reasoning capabilities of 23 state-of-the-art LLMs. This comprehensive evaluation enables us to establish a clear ranking of their reasoning performance. Further analysis across these difficulty levels reveals that many current LLMs outperform university-level students on easier reasoning tasks, while reasoning LLMs demonstrate significant advantages on more challenging tasks, achieving comparable or even superior performance compared to university-level students.
To investigate the impact of data contamination, we simulate scenarios by fine-tuning LLMs using golden trajectories generated by an optimal search algorithm. Our findings indicate that such contaminated LLMs exhibit poor generalization to out-of-domain tasks and tasks with varying difficulty levels. This underscores \method's robustness against dataset contamination, as it continuously updates tasks across new domains.
Additionally, our statistical analyses of \method reveal an intriguing pattern: domains with similar entity-relation graph topologies consistently exhibit comparable reasoning performance across LLMs. This suggests that employing a diverse range of domains is crucial to thoroughly assess various facets of reasoning ability and to accurately evaluate model generalization.
Overall, our benchmark results strongly indicate that LLMs genuinely possess reasoning abilities, as the dynamically generated tasks are entirely unseen during training.

\section*{Results}\label{sec:results}

\subsection*{\method benchmark}\label{subsec:benchmark}

We introduce a complex reasoning game designed to evaluate the reasoning capabilities of LLMs based on their gameplay trajectories. A single game instance is defined by the following components:

\begin{itemize}
    
    \item Truth Set ($\mathcal T = \{t_i\}_{i=1}^{N^{\text{truth}}}$): a finite and countable set containing $N^{\text{truth}}$ possible truths.
    \item Action Set ($\mathcal A = \{a_i\}_{i=1}^{N^{\text{action}}}$): a finite and countable set containing $N^{\text{action}}$ possible actions.
    \item Outcomes ($\mathcal{O}: a\mapsto o_a$): a mapping that associates each action $a \in \mathcal A$ with its outcome $o_a \in \mathcal O_a$, where $\mathcal O_a$ is the set of all possible outcomes for action $a$.
    \item Knowledge Book ($\mathcal K$): a document providing definitions of $\mathcal T, \mathcal A, \mathcal{O}$, and detailing how each truth $t \in \mathcal T$ can be ruled out from observed outcomes in natural language.
\end{itemize}

\begin{figure}
    \centering
    \includegraphics[width=0.95\linewidth]{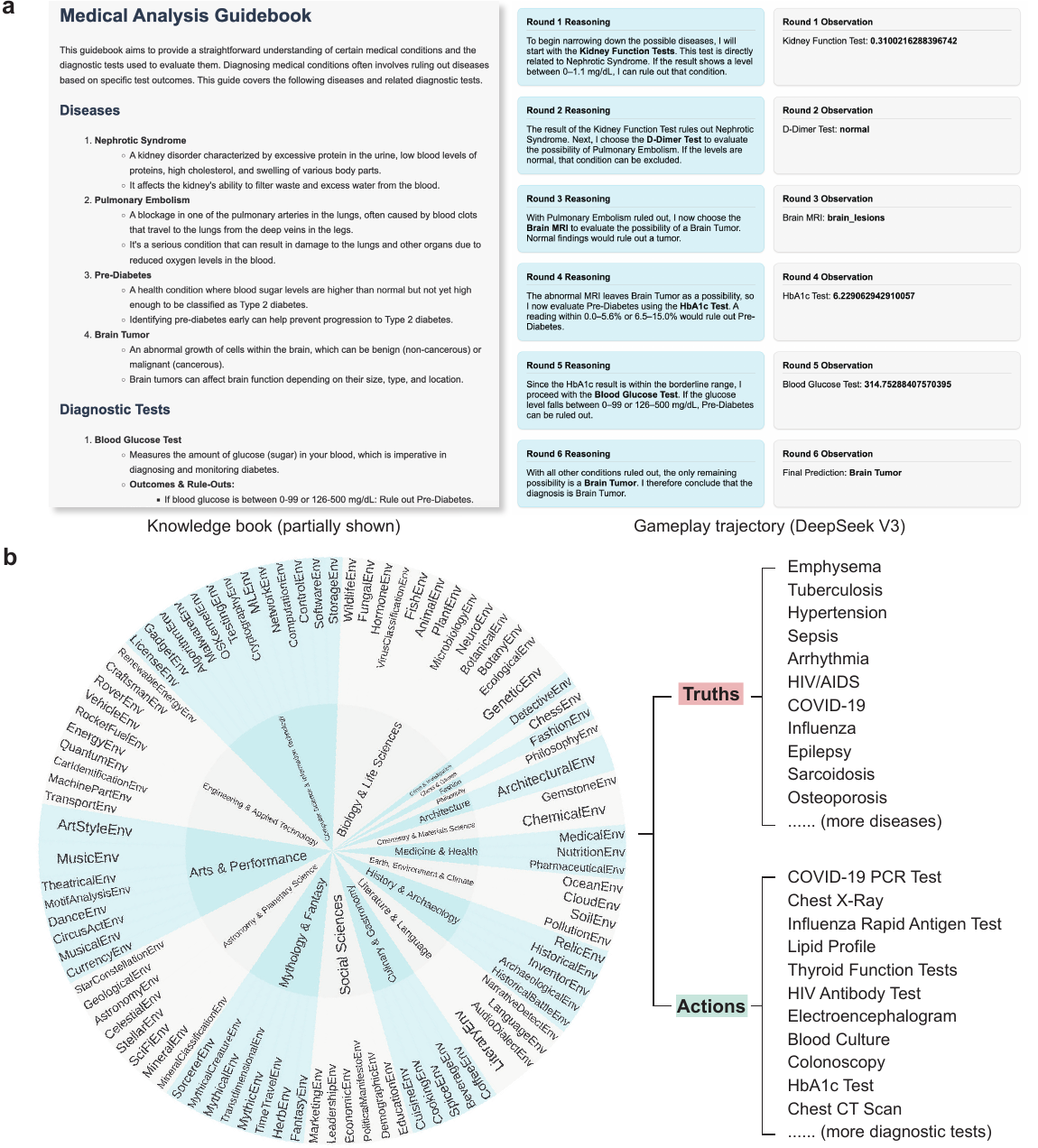}
    \caption{Overview of \method tasks. \textbf{a. An example of the complex reasoning game.} In this game, the player is presented with a list of potential ``truths'', available ``actions'', and a knowledge guidebook for a specific scenario. In the illustrated case of a diagnostic test scenario, the ``truths'' represent diseases, and the ``actions'' correspond to diagnostic tests. During each round, the player selects one action, observes its ``outcome'', and uses the information to eliminate invalid truths. The objective is to identify the single valid truth using the fewest possible actions. \textbf{b. The generated tasks in \method.} This study employs an automated pipeline to generate 100 exemplar task environments across 18 topic categories. Each environment includes approximately 50 truths and 30 actions. The figure shows part of the truths and actions from the Medical environment, which corresponds to the scenario depicted in panel a.}
    \label{fig:1}
\end{figure}

At the start of the game, a valid truth $t^\star \in \mathcal T$ is initialized, with all other truths in $\mathcal T$ marked as invalid. Outcomes are then generated to be consistent with the valid truth $t^\star$. In each round, the player selects an action $a \in \mathcal A$, and the game reveals the corresponding outcome $o_a$. The objective of the game is to accurately identify the valid truth $t^\star$ using the fewest possible actions. For instance, as illustrated in Fig.~\ref{fig:1}a, consider a medical analysis game. The player is provided with a medical analysis guidebook ($\mathcal K$), which details the relationship between various diseases ($\mathcal T$) and diagnostic tests ($\mathcal A$). In each round, the player can choose a diagnostic test ($a$) to observe its corresponding outcome ($o_a$) or predict the disease for the patient. The goal is to minimize the number of tests required to make a correct disease prediction.

Solving this game requires sophisticated reasoning due to its complexity and partial observability. Assuming all truths have an equal probability of being the valid truth, the expected minimum number of actions required to identify  $t^\star$ can be computed (Methods). Optimal gameplay involves selecting actions that minimize the expected number of steps in subsequent rounds. Calculating the expected minimum steps requires a recursive search process, closely tied to the player’s planning horizon. Additionally, as the game is partially observable (the outcome is unknown to the player in advance), the player must dynamically adjust their strategy based on observed outcomes. This interplay between planning and observation mirrors the strategic depth of games like chess, making the game an effective tool for assessing complex reasoning abilities in LLMs.

\method offers several advantages as a benchmark for evaluating LLMs. First, it provides ground truth, enabling automatic evaluation that is both efficient and objective. Second, \method is scalable. We have developed a flexible framework for automatic task generation (Fig.\ref{fig:2}), demonstrating that tasks within this benchmark can be easily created and updated. In this paper, we automatically generate 5,200 tasks in 100 different domains to evaluate the state-of-the-art open-sourced LLMs from a wide range of configurations (Fig.\ref{fig:4}).
Furthermore, \method supports adjustable difficulty levels (Fig.\ref{fig:3}) and our results reveal that it effectively resists saturation, as evidenced by a  performance gap between LLMs and the oracle search engine (Fig.\ref{fig:3}b). Lastly, this generative design mitigates data contamination issues, ensuring a reliable evaluation process (Fig.\ref{fig:5}).

\subsection*{A scalable framework for task generation}

\begin{figure}
    \centering
    \includegraphics[width=1.0\linewidth]{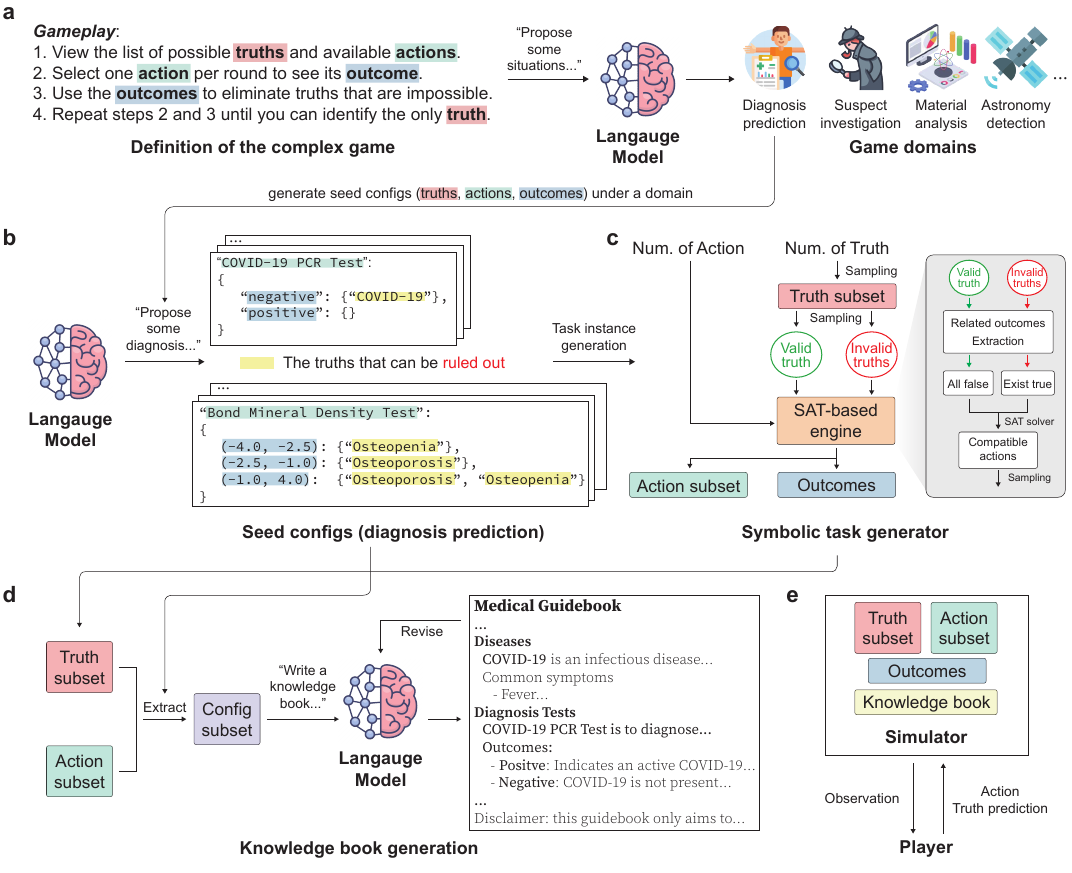}
    \caption{The construction of \method benchmark consists of five stages: \textbf{a. Domain proposal}. A capable LLM is prompted to propose various scenarios for the complex game based on its definition. These scenarios, referred to as domains, are collected. \textbf{b. Seed config generation.} The LLM is further prompted to generate foundational elements for each domain, including truths, actions, and their corresponding outcomes. These outcomes are designed to rule out certain truths. \textbf{c. Task instance generation.} To create a specific task instance, the sizes of its candidate truth set and action set are first determined. A subset of truths is then sampled from the universal truth set, with one selected as valid while the others are treated as invalid. The generation of compatible actions and outcomes is modeled as a satisfiability (SAT) problem. An SAT-based engine is employed to sample the action subset and generate outcomes. This process involves extracting related outcomes for each truth, assigning logical values based on validity, and using a SAT solver to produce a viable solution. \textbf{d. Knowledge book generation.} Once a task instance is generated, an LLM is tasked with writing a knowledge book and revising it if any error detected. This book translates the outcome configurations associated with the sampled truth and action subsets into detailed natural language descriptions. \textbf{e. Evaluation.} In each round, the player takes actions or makes truth prediction, and a simulator provides observations for the action based on the outcomes of the task (which is unseen to the player). The goal is to achieve accurate truth prediction while minimizing the number of actions taken.}
    \label{fig:2}
\end{figure}

The tasks in \method are constructed through a multi-stage pipeline (Fig.\ref{fig:2}). Each stage uses the generative capabilities of LLMs and symbolic engines to create game scenarios from scratch. The first stage (Fig.\ref{fig:2}a) applies LLM to propose a set of diverse domains, each representing a distinct scenario within the game. These could include predicting a patient’s disease, identifying an unknown material through chemical analysis, or diagnosing student learning gaps. An LLM is prompted with a general template (Extended Data Figs.1-2) to propose candidate domains aligned with the game’s definition. These domains define the thematic structure of the tasks and reflect real-world reasoning contexts. Each domain maps to a unique graph structure of truths and actions (Fig.\ref{fig:6}d), shaping how reasoning unfolds and supporting a broad evaluation of reasoning skills. The use of diverse domains also reduces the effect of dataset contamination (Fig.\ref{fig:5}).

After selecting the domains, the LLM generates seed configurations for each one: \emph{truths} (domain-specific propositions) and \emph{actions} with their outcomes (Fig.\ref{fig:2}b, Extended Data Fig.3). The outcomes are crafted to eliminate certain truths, requiring players to reason through the space of possibilities to identify the correct ones.

To instantiate a concrete task within a chosen domain (Fig.\ref{fig:2}c), we first determine the size of the candidate truth set and the action set. From a universal pool of truths, a subset is sampled, designating one as valid and treating the others as invalid. The generation of compatible actions and corresponding outcomes is formalized as a satisfiability (SAT) problem. An SAT-based engine is then employed to pick the action subset and produce outcomes consistent with the valid truth. The process involves extracting related outcomes, assigning logical values based on validity, and using an SAT solver to ensure a logically coherent configuration.

Once a task instance is created, an LLM is tasked with writing a knowledge book (Fig.\ref{fig:2}d, Extended Data Fig.4), which provides a detailed, natural language description of the selected truths and actions. This knowledge book functions as a narrative complement to the raw configuration, converting logical outcomes and constraints into clear, scenario-specific documentation. However, the initial version of the knowledge book may contain logical inaccuracies or unclear language. Therefore, it is crucial to verify its precision and clarity. To do so, we randomly sample a significant number of knowledge books and identify recurring error patterns. The LLM is then prompted to detect any errors in the current knowledge book (Extended Data Fig.5). If errors are identified, the original version is revised, and a new, refined knowledge book is generated. We also perform a preliminary validation of the knowledge book refinement by testing LLMs on versions of the knowledge book with and without refinement for \method. The results show improved performance with the refined knowledge book (Supplementary).

With the task instance and its corresponding knowledge book finalized, an evaluation process commences (Fig.\ref{fig:2}e). The player—unaware of the underlying task configuration—interacts with the environment by taking actions or making predictions about the underlying truth. A simulator responds with outcomes generated by the SAT engine. The player's objective is to correctly identify the valid truth with minimal action counts.

\subsection*{Benchmarking State-of-the-Art LLMs}

\begin{figure}
    \centering
    \includegraphics[width=1.0\linewidth]{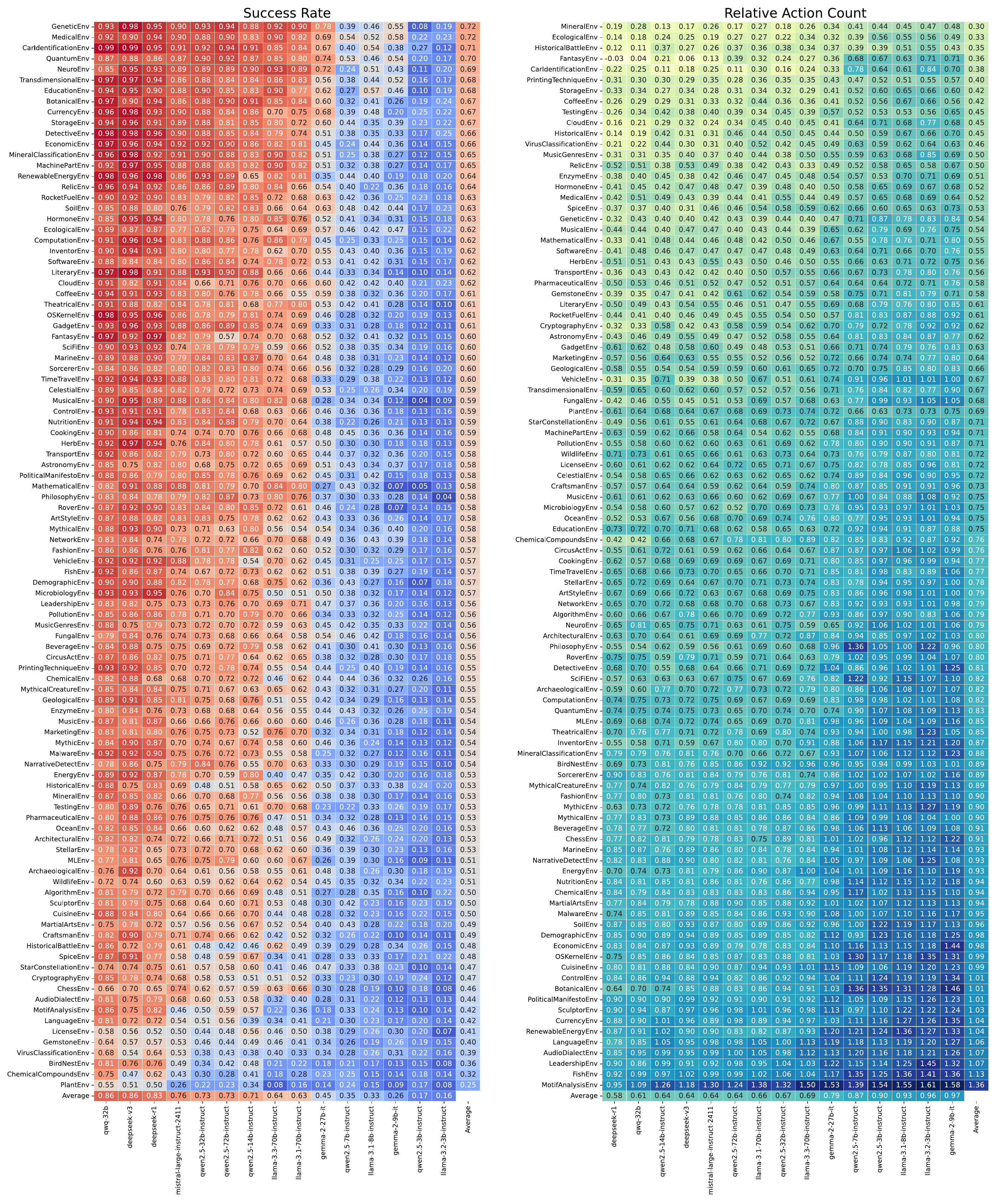}
    \caption{Benchmark results for 100 domains in the Easy setting (\#Truths=4, \#Actions=6) using \method for open-sourced Large Language Models (LLMs). \textbf{Left panel:} Success rates of LLMs, ranked from highest to lowest from left to right. \textbf{Right panel:} Relative action counts of LLMs. Domains are ranked from top to bottom based on the average metrics across LLMs.}
    \label{fig:3}
\end{figure}

\begin{figure}
    \centering
    \includegraphics[width=1.0\linewidth]{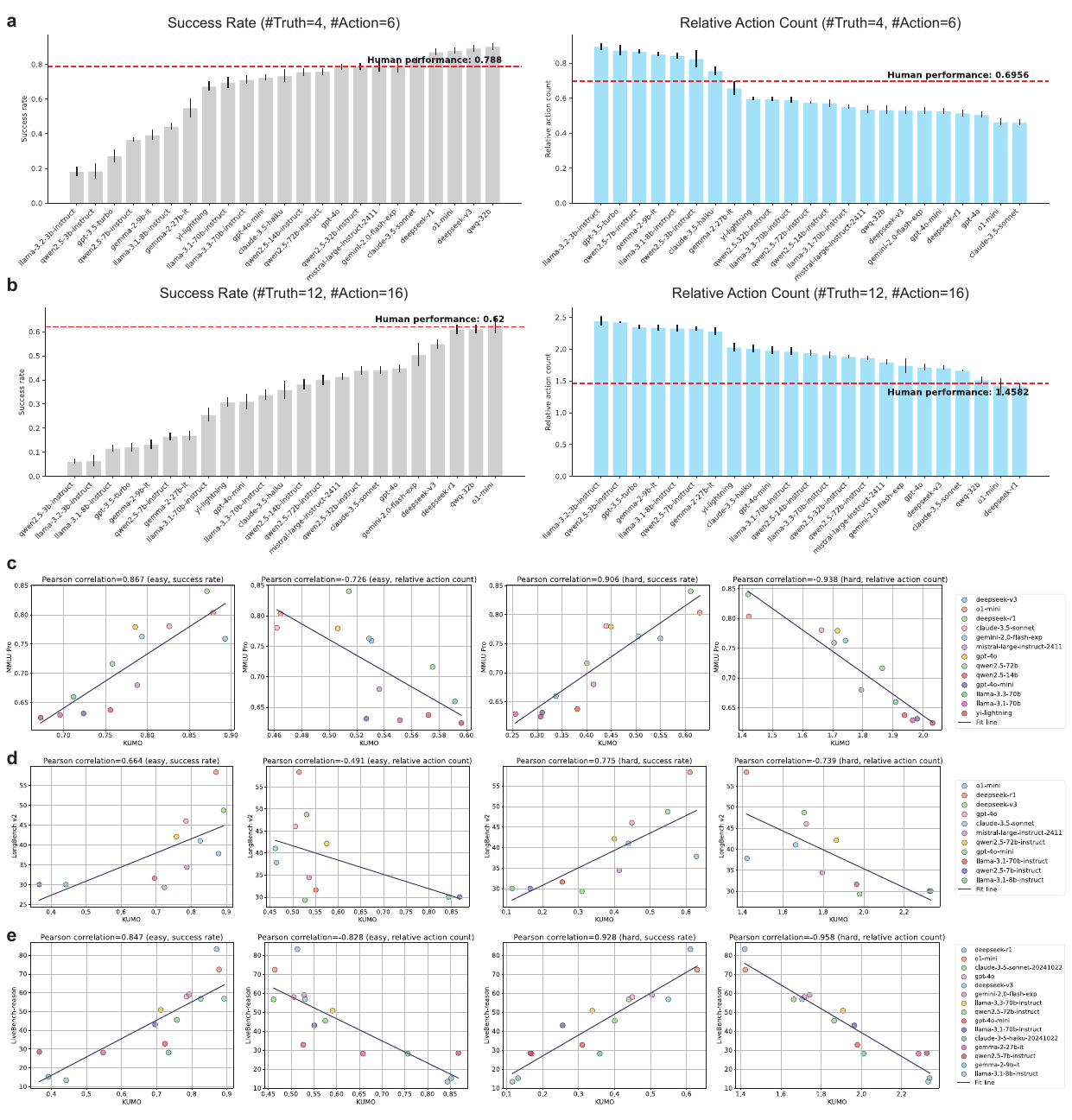}
    \caption{Benchmarking Large Language Models (LLMs) on \method and correlation with other LLM benchmarks. We evaluate 23 state-of-the-art LLMs varying in parameter counts, architectures, and organizational origins across five environments: MedicalEnv, ChemicalEnv, EducationEnv, FantasyEnv, and MusicEnv. Each environment has two difficulty levels: Easy (\#Truths=4, \#Actions=6) and Hard (\#Truths=12, \#Actions=16). \textbf{a.} Success rate and relative action count metrics for the Easy setting. \textbf{b.} Success rate and relative action count metrics for the Hard setting. Pearson Correlation of LLM performance between \method and \textbf{c.} MMLU-Pro benchmark, \textbf{d.} LongBench-V2 benchmark, and \textbf{e.} LiveBench-Reason benchmark.}
    \label{fig:4}
\end{figure}

We generate 100 domains (Fig.\ref{fig:1}b) and conduct a comprehensive evaluation of 23 state-of-the-art LLMs across two difficulty levels: \textit{Easy} (\#Truths = 4, \#Actions = 6) and \textit{Hard} (\#Truths = 12, \#Actions = 16). Due to cost considerations, we evaluate 15 open-source LLMs on the Easy setting across all 100 domains (Fig.\ref{fig:3}), and evaluate all 23 LLMs on both Easy and Hard settings within a subset of 5 domains ({MedicalEnv}, {ChemicalEnv}, {MusicEnv}, {EducationEnv}, and {FantasyEnv}; Fig.\ref{fig:4}). We assess reasoning performance using two metrics: (1) \textbf{success rate}, defined as the percentage of trajectories that correctly predict the target truth; and (2) \textbf{relative action count}, which quantifies the deviation between the number of actions taken and the optimal number required (see Methods). A higher success rate reflects greater validity in reasoning, whereas a lower relative action count indicates more efficient reasoning trajectories. For each domain and difficulty level, we generate 50 task instances and evaluate each LLM over 5 runs per instance.

We first observe that among the evaluated LLMs, three models---{QwQ-32B}, {DeepSeek-R1}, and {o1-mini}---are reasoning LLMs, meaning they generate reasoning thoughts before producing the answer. The remaining models are instruction-tuned LLMs. From experiments conducted across 100 domains, we find that powerful non-reasoning LLMs (e.g., {DeepSeek-V3}) are capable of solving these relatively simple reasoning tasks, achieving a higher success rate (0.86) compared to its reasoning counterpart LLM ({DeepSeek-R1}, at 0.83). Analyzing the trajectories, we observe that reasoning-scaled models tend to overthink, which can lead to incorrect predictions in easy setting. When comparing relative action counts, reasoning LLMs significantly outperform instruction-tuned ones, indicating that they are better at identifying efficient reasoning paths. Interestingly, we also find that larger model size does not necessarily correlate with better reasoning performance, as seen in comparisons between the {Qwen2.5} and {LLaMA} series.

Similar observations are drawn from the five-domain experiments (Figs.\ref{fig:4}a-b). In the hard setting, reasoning LLMs demonstrate remarkable performance compared to instruction-tuned LLMs—the three reasoning LLMs consistently rank first, with a significant performance gap. Interestingly, when compared to human performance (university students), LLMs outperform humans in the easy reasoning setting, achieving both a higher success rate and a lower relative action count. However, in the difficult setting, only the reasoning LLMs reach performance levels comparable to those of humans. These findings suggest that current LLMs are capable of replacing humans in many simple reasoning tasks. For more complex tasks, however, human reasoning remains more reliable—though reasoning LLMs show strong potential to surpass human performance in the future.

Since \method is a synthetic benchmark, we validate its relevance to real-world applications by computing the Pearson correlation between LLM performance on \method and on other established benchmarks (Figs.~\ref{fig:4}c–e). The selected benchmarks MMLU-Pro, LiveBench-Reason, and LongBench V2, are all recently published and thus likely unaffected by training data contamination. Overall, we observe a clear positive correlation for success rate and a negative correlation for relative action count when compared to these benchmarks. Notably, the correlations with MMLU-Pro and LiveBench-Reason are significantly higher than with LongBench V2. This is likely because LongBench V2 emphasizes long-context understanding, while \method focuses more on reasoning. We also find that correlations are substantially higher under the hard setting of \method than the easy setting (within each benchmark and both the metric), suggesting that the selected benchmarks are themselves challenging. In fact, correlation with \method may serve as a proxy for assessing the difficulty of reasoning benchmarks. Given that correlations exceed 0.9 in some cases, we conclude that \method is a reliable and scalable benchmark for evaluating the reasoning abilities of LLMs, with the added benefit of being contamination-free.




\subsection*{\method resists overfitting}

\begin{figure}
    \centering
    \includegraphics[width=1.0\linewidth]{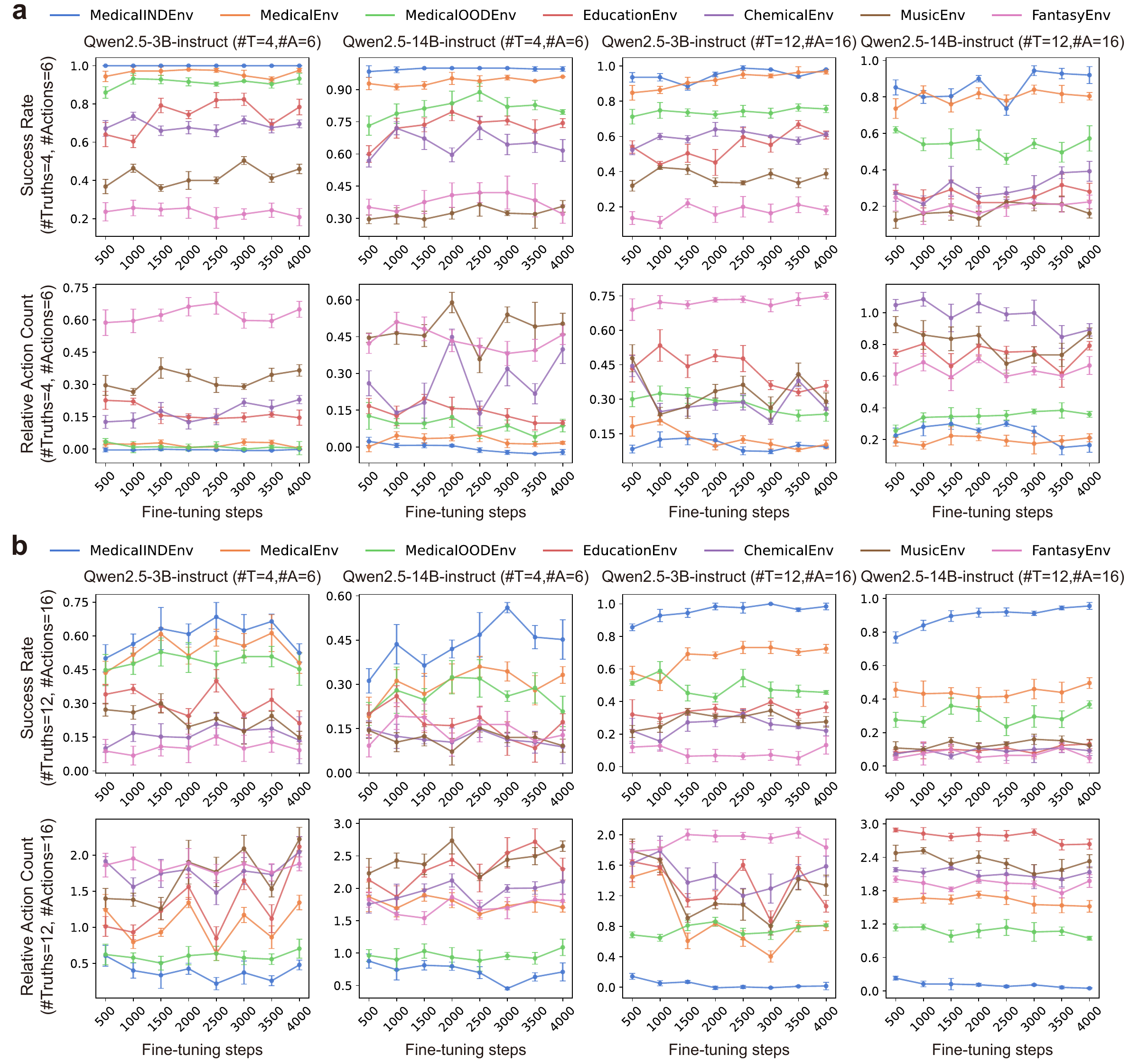}
    \caption{Performance of Large Language Models (LLMs) fine-tuned on golden trajectories. The MedicalEnv environment is divided into MedicalINDEnv (in-distribution) and MedicalOODEnv (out-of-distribution), each with distinct connection components. Two LLMs, Qwen2.5-0.5B-Instruct and Qwen2.5-3B-Instruct, are fine-tuned on golden trajectories within MedicalINDEnv under Easy (\#Truths=4, \#Actions=6) and Hard (\#Truths=12, \#Actions=16) settings. \textbf{a.} Success rate and relative action count metrics for the Easy setting. \textbf{b.} Success rate and relative action count metrics for the Hard setting. Fine-tuned LLMs exhibit strong in-distribution (IND) generalization but experience severe performance degradation for out-of-domain (OOD) generalization and difficulty transitions (Easy to Hard / Hard to Easy). This demonstrates the benchmark's resistance to overfitting through diverse setting generation.}
    \label{fig:5}
\end{figure}

Since the code for \method will be publicly available, we aim to maintain a faithful \method leaderboard, with domain updates occurring every two months. To address potential overfitting—where models might exploit our task generation pipeline to synthesize a large number of task instances—we examine whether LLMs that overfit on a single domain within one round can still perform well on other domains in subsequent rounds (i.e., the following two months). We fine-tune LLMs on golden trajectories generated by the optimal search algorithm (see {Methods}) in one domain and evaluate their performance both in-domain and out-of-domain. For this, we split {MedicalEnv} into two disjoint subdomains based on connection anlaysis (Methods)—{MedicalINDEnv} and {MedicalOODEnv}—to simulate an out-of-distribution but in-domain setup. The remaining domains ({EducationEnv}, {ChemicalEnv}, {MusicEnv}, {FantasyEnv}) serve as additional out-of-domain evaluation environments. We also evaluate out-of-difficulty generalization, testing whether LLMs fine-tuned on one difficulty level can generalize to another (easy to hard and hard to easy).

The experimental results (Fig.\ref{fig:5}) demonstrate that fine-tuned LLMs achieve strong performance on in-distribution generalization, with all models performing best in {MedicalINDEnv}. They also show good in-domain and hard-to-easy generalization, as evidenced by good results in {MedicalOODEnv} and {MedicalEnv}. However, out-of-domain and easy-to-hard generalization remain challenging. Notably, performance varies significantly across domains: for instance, fine-tuned LLMs on {FantasyEnv} perform at near-random levels. These findings suggest that \method presents a substantial challenge for both out-of-domain generalization and difficulty-based generalization, effectively mitigating hacking to unseen domains and curbing difficulty saturation.

\subsection*{Statistical study of \method}

\begin{figure}
    \centering
    \includegraphics[width=\linewidth]{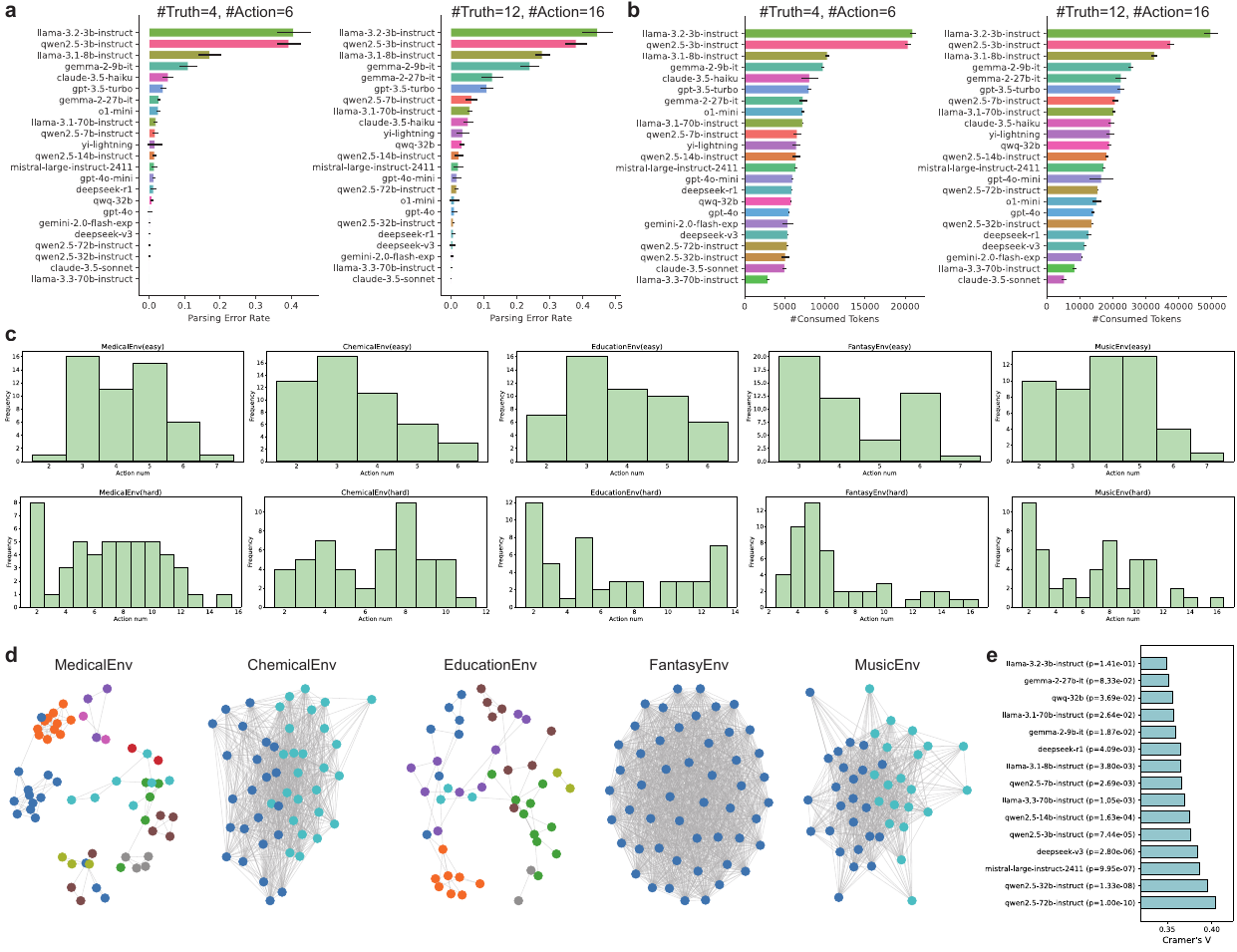}
    \caption{We use the results from benchmarking experiments to conduct a statistical analysis of \method. \textbf{a.} Parsing error rates across LLMs in easy and hard settings, averaged over 5 domains. \textbf{b.} Token consumption (input + output) across LLMs in easy and hard settings, averaged over 5 domains. \textbf{c.} Raw action counts for the optimal search algorithm across 5 domains. \textbf{d.} Graph topology for 5 domains: nodes represent truths, edges denote actions connecting truths as possible outcomes, and colors indicate Louvain community membership. \textbf{e.} Cramér’s V between LLM performance and domain graph topology, computed over all 100 domains.}
    \label{fig:6}
\end{figure}

In our analysis of parsing errors (the percentage of generated text that cannot be parsed into truths or actions) and token consumption (both the number of input and output tokens) for \method (Figs.\ref{fig:6}a-b), we observe that certain LLMs—such as LLaMA-3.2-3B, Qwen2.5-3B-instruct, LLaMA-3.1-8B-instruct, and Gemma-2-9B-it—consume significantly more tokens than others while also exhibiting high parsing error rates. These models frequently fail to follow instructions and tend to generate excessively long and irrelevant outputs, and also achieve very bad performance in even easy setting of \method (Fig.\ref{fig:3}). This shows that \method is challenging for small models. 

Among the more capable models that demonstrate nontrivial performance on \method, token consumption in the easy setting ranges from 2,937.00 tokens (LLaMA-3.3-70B-instruct) to 8,121.88 tokens (Claude-3.5-Haiku), with an average of 6,126.35 tokens. In the hard setting, consumption spans from 5,395.78 tokens (Claude-3.5-Sonnet) to 22,605.71 tokens (Gemma-2-27B-it), with an average of 16,002.93 tokens. Overall, the number of consumed tokens scales almost proportionally with the size of the action space, indicating a near-linear expansion of the search space as the number of actions increases.

In the analysis of optimal raw action count (Fig.\ref{fig:6}c), we find that in the easy setting, most optimal action counts fall within the range of 3 to 5, with an expected value of 3.92. In the hard setting, optimal action counts are generally much smaller relative to the size of the action space, with two actions being the most frequent choice and an expected value of 6.69. This suggests that even in more complex scenarios, effective strategies exist for solving the task, underscoring the importance of strategic action selection. Moreover, the distribution of optimal action counts varies across domains, reflecting the diversity and domain-specific nature of the task.

In our analysis of domain graph topology and its correlation with reasoning performance (Figs.\ref{fig:6}d-e), all models—except {LLaMA-3.2-3B-instruct} and {Gemma-2-27B-it}—exhibit statistically significant correlations (as measured by p-values) between the internal logical graph structures of tasks in \method and their corresponding success rates. This indicates that a model’s ability to solve a particular task in our benchmark is meaningfully influenced by the logical structure of the domain itself. One possible explanation is that similar domains require similar reasoning capabilities, which leads to correlated performance across them (Fig.\ref{fig:5}). These findings highlight the importance of evaluating reasoning performance across diverse domains, as each domain embodies distinct logical patterns.

\section*{Discussion}\label{sec:discussion}

When synthesizing tasks in \method, the occurrence of some counterfactual instances is unavoidable. However, addressing counterfactuality is beyond the primary scope of \method, which specifically targets the evaluation of LLM reasoning capabilities independently from their internal knowledge—hence the inclusion of a comprehensive external knowledge source (the knowledge book) during reasoning. Counterfactual content can potentially mislead LLMs due to conflicts between their implicit world knowledge and the externally provided knowledge book. Despite this, \method can readily be adapted into a counterfactual reasoning benchmark by exclusively generating counterfactual examples. Similar adaptations could extend \method to other reasoning dimensions, such as long-context reasoning (utilizing a longer knowledge book), probabilistic reasoning (implementing probabilistic exclusion rules), or multi-truth reasoning (allowing multiple valid truths). Overall, \method demonstrates significant potential as a versatile generative benchmark framework adaptable to diverse research objectives.

In real-world tasks, reasoning is inherently intertwined with knowledge acquisition, which typically involves leveraging web-scale data during training or employing retrieval-based augmentation methods (e.g., RAG). Consequently, reasoning-oriented applications encompass multiple stages: knowledge acquisition, problem analysis, knowledge retrieval, and logical reasoning. This paper aims to disentangle logical reasoning from these interconnected abilities, clearly illustrating that LLMs can perform on par with humans in challenging reasoning tasks. Nevertheless, the efficiency and accuracy of these models still have room for improvement to achieve optimal performance. Future research should further investigate knowledge-related capabilities to build a comprehensive understanding of logical reasoning in practical, real-world contexts.

\section*{Methods}\label{sec:methods}
\subsection*{Language models} \label{subsec:LLM}


In this study, we benchmark a diverse set of LLMs to assess their reasoning performance across a range of tasks using \method. A total of 23 models were included in the analysis, comprising the Qwen2.5 series (3B, 7B, 14B, 32B, 72B), LLaMA series (3.2-3B, 3.1-8B, 3.1-70B, 3.3-70B), Gemma-2 series (9B, 27B), Claude3.5 series (Haiku, Sonnet), GPT series (GPT-3.5-turbo, GPT-4o-mini, GPT-4o, o1-mini), Deepseek series (V2.5, V3, R1), QwQ-32B, Gemini-2.0-flash-exp, and Yi-lightning. The models were grouped into two broad categories: open-source models (LLaMA, Qwen, Deepseek, Gemma variants) and closed-source models (including those from OpenAI, Anthropic, 01.AI, and Google DeepMind). The parameter counts of these models span from a few billion to several hundred billion, allowing us to explore the performance of both lightweight models suited for resource-constrained environments and cutting-edge models optimized for high-performance tasks.
In addition to evaluation, we use OpenAI’s o1 for domain proposal and seed configuration generation, and apply GPT-4o for knowledge book creation (Fig.\ref{fig:2}) as it is more cost-efficient. We also conduct an ablation study to verify there is no significant bias between configurations and knowledge books generated by different LLMs (Supplementary).

We call the LLMs via APIs from the official websites of close-source models and the Deepseek series. For other open-sourced models, we serve them on our own machine using the VLLM framework. Both API-based and VLLM-served models use the default temperature setting for LLM decoding.

\subsection*{Automatic domain proposal and seed configuration generation}

Users can interact with \method via a Jupyter notebook to propose new domains, generate seed configurations, and create a registered game environment class. Domain proposals utilize LLMs to generate natural language metadata, including descriptions of reasoning goals, truths, and domain-specific actions. For selected domains, the metadata is integrated into a prompt template (Extended Data Fig.2), forming a domain-specific prompt used for seed configuration generation (Extended Data Fig.3). The LLM then produces domain-specific seed configurations, which include a universal truth set $\mathcal T^{\text{univ}}$, a universal action set $\mathcal A^{\text{univ}}$, outcomes $\mathcal O$, and a symbolic version of the knowledge book $\mathcal K^{\text{symb}}$. When creating a task instance under a specific domain, the truth set $\mathcal T$ and action set $\mathcal A$ will be further sampled from $\mathcal T^{\text{univ}}$ and $\mathcal A^{\text{univ}}$, and $\mathcal K^{\text{symb}}$ will be rewritten in nature language by LLM. To handle potential truncation issues, the LLM is iteratively queried to verify the completeness of the output. Generation continues from truncation points as necessary, up to a predefined retry limit (currently set to 3 attempts by default). Generated configurations are parsed into Python data structures, stored, and validated to ensure no truth is universally excluded by available actions. Invalid configurations are regenerated as needed. Finally, another prompt template (Extended Data Fig.4) is used with an LLM to generate a Python file defining the registered environment class.

\subsection*{SAT-based task generation engine}
\label{sec:sat_based_task_generation}
\noindent
The SAT-based task generation engine systematically produces consistent and diverse task instances based on a predefined domain (e.g., MedicalEnv, ChemicalEnv) characterized by a universal truth set $\mathcal T^{\text{univ}}$, a universal action set $\mathcal A^{\text{univ}}$, and an outcome mapping $\mathcal{O}$ and the symbolic version of the knowledge book $\mathcal K^{\text{symb}}$. The final truth set $\mathcal T$ and action set $\mathcal A$ for a generated task will be a subset of the universal truth set $\mathcal T^{\text{univ}}$ and action set $\mathcal A^{\text{univ}}$. The generation process is governed by three primary parameters: the total number of truths $N^{\text{truth}}$, the number of actions $N^{\text{action}}$, and the number of valid truths $N^{\text{valid}}$. In this study, we set $N^{\text{valid}} = 1$, although this parameter can generally take any positive integer value less than $N^{\text{truth}}$.

Initially, a subset of truths $\mathcal T^{\text{sub}}$ consisting of $\mathcal N^{\text{truth}}$ elements is randomly sampled without replacement from the universal truth set $\mathcal T^{\text{univ}}$. Within $\mathcal T^{\text{sub}}$, exactly $N^{\text{valid}}$ truths ($\mathcal T^{\text{valid}}$) are randomly designated as valid, representing conditions that observations cannot contradict. For every action $a \in \mathcal A^{\text{univ}}$ and each associated outcome $o_a\in \mathcal O_a$, consistency with respect to the selected truths is evaluated. An outcome $o_a$ is considered contradictory if it excludes any truth from $\mathcal T^{\text{valid}}$; otherwise, it is deemed valid. The engine utilizes a SAT solver to select appropriate combinations of actions and outcomes subject to specific constraints: each action may have at most one outcome selected (unique state per action constraint), the total number of selected actions must not exceed $N^{\text{action}}$ (action limit constraint), and each invalid truth (truths in $\mathcal T^{\text{sub}}$ but not in $\mathcal T^{\text{valid}}$) must be excluded by at least one selected outcome (invalid truth exclusion constraint). If fewer than $N^{\text{action}}$ actions are selected initially, additional actions are included, first choosing unused related (non-contradictory) actions, and subsequently, if required, irrelevant actions to meet the action count requirement. This procedure is iterated to generate the desired number of task instances, each of which is checked against existing instances to prevent duplication of truths, actions, and observed outcomes. The pseudo code is in Supplementary.

\vspace{2em}

\subsection*{Optimal search algorithm}

The optimal search algorithm leverages recursion to determine the minimal expected steps and the corresponding optimal action for the given truth space, $\mathcal T^{\text{current}}$, and action space, $\mathcal A^{\text{current}}$, which are set as the vanilla truth set $\mathcal T$ and action set $\mathcal A$. The recursion continues until specific base conditions are met. Recursion terminates if either the current truth space size is one or fewer truths remain, or if the action space is empty. Another termination condition occurs if at least one truth in the current set is unrelated to any remaining actions (note that we set $N^{\text{valid}}=1$). In both scenarios, the expected number of steps required, denoted as $\mathbb{E}[S]$, is one, signifying no further actions are necessary and only need to output the determined truth.

If the base conditions are not satisfied, the algorithm constructs a binary bitmask representation, $B$, to encode the current state efficiently. This bitmask assigns unique indices to each truth (indexed $0$ to $|\mathcal T|-1$) and each action (indexed $|\mathcal T|$ to $|\mathcal T|+|\mathcal A|-1$), computing the mask as:
\begin{equation}
B = \sum_{t \in \mathcal T^{\text{current}}} 2^{\text{idx}(t)} + \sum_{a \in \mathcal A^{\text{current}}} 2^{\text{idx}(a)}.
\end{equation}

The algorithm then checks if this bitmask $B$ exists in a memoization table, $\text{BestActionDict}$. If a match is found, it retrieves the stored values of expected steps $\mathbb{E}[S]$ and the optimal action $a^*$ directly from the table, avoiding redundant computation.

When the bitmask is not present in $\text{BestActionDict}$, the algorithm proceeds with recursive computation, initializing the minimal expected steps $\mathbb{E}[S]_{\text{min}}$ to infinity and the best action $a^*$ to undefined. It iteratively evaluates each possible action $a$ within the current action space. For each evaluated action, the algorithm adjusts the subsequent action space by excluding the selected action, thus defining $\mathcal A^{\text{next}} = \mathcal A^{\text{current}} \setminus \{a\}$.

To quantify state transitions, the algorithm calculates state probabilities based on the uniform assumption across truths in the current space. Each action $a$ leads to different outcomes $o_a \in \mathcal O_a$, with each outcome excluding certain truths and lead to a new truth set $\mathcal T_{o_a}\subset \mathcal T^{\text{current}}$. Outcome probabilities $P(o_a)$ are calculated using a weighting factor $W(o_a)$, defined as the size of the current truth space $\mathcal T_{o_a}$. The probability for each state is normalized accordingly:
\begin{equation}
W(o_a)=|\mathcal T_{o_a}|,\quad P(o_a)=\frac{W(o_a)}{\sum_{o_a'\in \mathcal O_a}W(o_a')+\epsilon}, \quad \forall o_a \in \mathcal O_a,    
\end{equation}

where $\epsilon$ prevents division by zero.

Subsequently, the algorithm recursively computes the cumulative expected steps for each action, $\mathbb{E}[S_a]$. For every result $o_a$ resulting from action $a$, it generates a reduced truth space $\mathcal{T}_{o_a}$ by excluding the truths removed by state $o_a$. The expected steps for these reduced spaces are calculated recursively, weighted by their probabilities, and summed to update the cumulative steps $\mathbb{E}[S_a]$. If at any point $\mathbb{E}[S_a]$ surpasses the current minimum $\mathbb{E}[S]_{\text{min}}$, the algorithm prunes further evaluation for action $a$. After evaluating all states for each action, if the newly computed $\mathbb{E}[S_a]$ is lower than the current minimum, the algorithm updates $\mathbb{E}[S]_{\text{min}}$ and designates the current action as the best choice, $a^*$. The resulting minimal expected steps and corresponding optimal action are stored in the memoization table $\text{BestActionDict}$ for future reference. Finally, the algorithm returns the optimal solution with the minimal expected number of steps, along with the optimal action $a^*$.

\subsection*{Evaluation metrics} \label{subsec:metric}

We evaluate the reasoning performance using two metrics: (1) \emph{Success Rate} measures whether the model correctly identifies the valid truth. It assigns a binary score, where 1 indicates correct identification, and 0 otherwise. The success rate is computed as:
\begin{equation}
    \text{Success Rate} = \frac{\text{Number of Correct Identifications}}{\text{Total Number of Tasks}}
\end{equation}
\emph{Relative Action Count} evaluates efficiency by comparing the number of actions taken by the model against an optimal baseline. Specifically, it measures the deviation between the model's actions and the optimal number of actions (determined through an optimal search strategy). It is computed as:
\begin{equation}
    \text{Relative Action Count} = \frac{\text{Model Action Count} - \text{Optimal Action Count}}{\text{Optimal Action Count}}
\end{equation}
A lower relative action count indicates closer alignment with optimal reasoning. We use relative action counts rather than absolute counts to normalize for task-specific differences, as tasks naturally vary in complexity and inherently require differing numbers of actions.

\subsection*{Design of the human experiment}

A total of 92 participants were recruited for the study, comprising 48 males and 44 females. The participants' ages ranged from 18 to 29 years, with 77.2\% falling within the 19--22 age group. They represented diverse academic backgrounds, spanning 47 different majors across 72 universities. Among the participants, 82 were undergraduate students, 8 were pursuing master's degrees, and 2 were pursuing doctoral degrees.

The experimental procedure was conducted on an online platform specifically designed for this study, developed using Streamlit. Participants logged in using an assigned User ID and password. Each time, a participant was randomly assigned a complete task set, consisting of 10 reasoning tasks. Half of the tasks were ``Easy'' (\#Truths=4) and the other half were ``Hard'' (\#Truths=12). Each task covered five domains evenly: MedicalEnv, EducationEnv, MusicEnv, FantasyEnv, and ChemicalEnv. A detailed English knowledge book was presented on the left side of the screen, containing the information needed to complete the reasoning tasks. On the right side, participants selected actions from a menu. After selecting an action, the system displayed an observation related to that action, which could be used to eliminate invalid truths (Fig.\ref{fig:1}a).

The objective for participants was to identify the only valid truth based on the observations and the knowledge book, while minimizing the number of actions taken. Performance was also evaluated using the same metrics as those applied to LLMs. The total earnings per task set is calculated as 
$25 + \text{success rate} \times 15 - \text{\#action count} \times 0.1$, 
with penalties for incorrect answers and excessive actions.
To ensure data quality, participant behavior was monitored throughout the study. Any participant who failed to meet the data-quality threshold (providing random answers in a very short amount of time) was excluded from the analysis, and their compensation was forfeited. For different trajectories on the same task set which all passed the quality test, one trajectory is randomly selected. In the end, we collected 500 high-quality game trajectories covering all task sets.

\subsection*{Domain graph visualization}
We use a domain graph to represent and analyze the internal graph structures between truths within a domain. The domain graph presented (Fig.\ref{fig:3}e) is constructed by connecting pairs of truths that co-occur within the same action's state mappings, thus forming edges between them. To uncover the underlying community structure, we apply the Louvain community detection algorithm~\citep{Blondel2008FastUO}. This algorithm optimizes \emph{modularity}, defined mathematically as:

\begin{equation}
Q = \frac{1}{2m}\sum_{ij}\left[A_{ij} - \frac{k_i k_j}{2m}\right]\delta(c_i, c_j)
\end{equation}

where \(A_{ij}\) represents the adjacency matrix of the graph, \(k_i\) and \(k_j\) are the degrees of nodes \(i\) and \(j\), \(m\) is the total number of edges, and \(\delta(c_i, c_j)\) is the Kronecker delta function, which equals 1 if nodes \(i\) and \(j\) belong to the same community \(c\), and 0 otherwise. The Louvain algorithm iteratively optimizes this modularity measure to partition the graph into densely interconnected clusters. Each detected community is assigned a distinct color in the visualization, enabling immediate identification of semantically related truth groupings.

\subsection*{Environment split via connection analysis}

To partition a predefined domain into two disjoint sub-domains, we use a connection-based method. Each action is associated with a set of related truths; truths linked to the same action are considered connected. We construct a truth graph where nodes represent truths and edges connect those sharing an action. Using Depth-First Search (DFS), we identify connected components—clusters of interrelated truths. These components are then alternately assigned to two disjoint sets to maintain balance: truth set $T_1$ and truth set $T_2$, ensuring that related truths stay together. Each action is then categorized: if all its related truths lie in $T_1$, it is assigned to $A_1$; otherwise, it goes to $A_2$. Outcomes are split accordingly, preserving consistency across the data.

\subsection*{Details of overfitting resistance experiment}

We perform supervised fine-tuning (SFT) on LLMs using golden trajectories generated by our optimal search algorithm to assess \method’s resistance to overfitting. We evaluate two LLMs of different scales: {Qwen2.5-3B-instruct} and {Qwen2.5-14B-instruct}.
To support generalization analysis, we partition the {MedicalEnv} environment into two sub-domains—{MedicalINDEnv} and {MedicalOODEnv}—using our connection analysis method. Training and validation datasets are constructed from {MedicalINDEnv} under two difficulty settings: Easy (\#Truths = 4, \#Actions = 6) and Hard (\#Truths = 12, \#Actions = 16). For each setting, we sample 100,000 task instances, reserving the first 50 for validation and the remaining 99,950 for training.
Golden trajectories for these instances are generated using the optimal search algorithm. Each sample begins with a system message describing the game configurations and a symbolic knowledge book $\mathcal K^{\text{symb}}$ (a Python dictionary capturing outcomes and rule-out information). Unlike in benchmarking experiments, this symbolic format is used to reduce the cost of LLM-generated knowledge books. During training, only the next-token prediction loss for the actions in the trajectory is included in backpropagation; losses from system and user messages are masked out.
We apply Low-Rank Adaptation (LoRA~\citep{Hu2021LoRALA}) with a rank of $r = 16$ and a scaling factor of $\alpha = 32$. Optimization is performed using the AdamW~\citep{Kingma2014AdamAM, loshchilov2017decoupled} optimizer with a learning rate of $2\times10^{-4}$, no weight decay, $\beta_1 = 0.9$, $\beta_2 = 0.999$, and $\epsilon = 1\times10^{-8}$. The learning schedule includes a linear warm-up over the first 3\% of training steps, followed by cosine decay over the remaining 97\%. We use gradient accumulation to achieve a global batch size of 8 across all GPUs. Training proceeds for 100 epochs with no early stopping.

\subsection*{Details of Chi-square test}
Leveraging synthetic reasoning tasks to construct \method offers a key advantage: precise knowledge of each task’s underlying logical graph structure. This enables a statistical investigation into how reasoning performance relates to domain graph properties, by analyzing the correlation between graph structures and model outcomes.

In our experimental setup, we extract two core data components for each evaluated LLM based on the 100-domain experiment (Fig.\ref{fig:3}): (1) task correctness, and (2) a simplified representation of the task's graph structure. For correctness, each model is evaluated on five trials per task across 100 domains (with 50 tasks per domain, totaling 5,000 tasks). A task is labeled as correct if the model succeeds in at least 3 out of 5 trials, using majority voting to determine the final binary label. For the graph representation, we begin by constructing the full bipartite truth-action graph for each task. We then derive a simplified structural signature by concatenating the sorted degree sequences of the truth nodes and action nodes. This signature is treated as a categorical variable denoting graph structure.

To assess the relationship between structure and performance, we conduct a Chi-square test of independence between graph structure categories and binary correctness labels for each LLM. The p-value for the test is given by:
\begin{equation}
    p = 1 - F_{\chi^2}(X^2; k)
\end{equation}
where \( F_{\chi^2} \) is the cumulative distribution function of the chi-square distribution with \( k \) degrees of freedom, and \( X^2 \) is the observed chi-square statistic.

To quantify the strength of association between graph structure and correctness, we compute Cramér’s V:
\begin{equation}
    V = \sqrt{\frac{X^2}{n \cdot \min\{r-1, c-1\}}}
\end{equation}
where \( n \) is the total number of observations, and \( r \) and \( c \) are the number of rows and columns in the contingency table, respectively. Larger values of Cramér’s V indicate stronger dependence between task structure and model performance.

\section*{Data availability}
The \method generated dataset and the evaluation results can be downloaded from our official Github repository \url{https://github.com/linhaowei1/kumo}. The generated datasets is also open-sourced at HuggingFace \url{https://huggingface.co/datasets/pkuHaowei/kumo-easy} and \url{https://huggingface.co/datasets/pkuHaowei/kumo-hard} for our five-domain experiment.

\section*{Code availability}
The code for \method, including the domain proposal, seed configuration generation, symbolic task generator, knowledge book generation, and game simulator, as well as the benchmarking of LLMs on \method, is available at \url{https://github.com/linhaowei1/kumo}.

\bibliography{sn-bibliography}

\newpage

\section{Supplementary figures}

\lstset{
    basicstyle=\ttfamily\small, 
    breaklines=true,              
    breakindent=0pt,              
    breakatwhitespace,            
    columns=fullflexible,         
    showstringspaces=false,       
    literate={
      {“}{{\lq}}1 {”}{{\rq}}1 {，}{{,}}1
    }
}

\begin{figure}[htbp]
    \centering
    \begin{tcolorbox}[
        title=\textbf{Examplar domain proposal},
        colback=SeaGreen!10!CornflowerBlue!10,
        colframe=RoyalPurple!55!Aquamarine!100!,
        breakable,
        fontupper=\sffamily 
    ]
    \begin{lstlisting}
{ 
    'Goal': 'Identify the being disrupting the fabric of space-time',
    'Truths': 'Traits of transdimensional entities',
    'Actions': 'Interaction experiments / Dimensional stability monitoring / Entity behavior analysis'
}
\end{lstlisting}
\end{tcolorbox}
\caption{Exemplar domain proposal: Identifying the entity disrupting the fabric of space-time (a highly fictional scenario generated by a language model).}
\label{supp:fig1}
\end{figure}

\newpage

\begin{figure}[H]
    \centering
    \begin{tcolorbox}[
        title=\textbf{Seed config generation template},
        colback=SeaGreen!10!CornflowerBlue!10,
        colframe=RoyalPurple!55!Aquamarine!100!,
        breakable,
        fontupper=\sffamily 
    ]
    \begin{lstlisting}
Fill in the missing values in the prompt below to create a configuration for a reasoning game.
(Note: You only need to create the prompt, not the configuration!)

**TRUTH**: Domain or topic (e.g., "Diseases").
**ACTION**: Main activity (e.g., "Diagnosis").
**GOAL**: Objective (e.g., "identify the disease").

# Prompt Template
Generate a configuration in Python for a {DOMAIN} reasoning game. The goal is to determine {GOAL} from observed test outcomes. Follow the format below.

Requirements:
1. Truths: List {TRUTH} values for {GOAL} (e.g., {TRUTH_EXAMPLE1}, {TRUTH_EXAMPLE2}, {TRUTH_EXAMPLE3}).
2. Actions: List {ACTION} values for {GOAL} (e.g., {ACTION_EXAMPLE1}, {ACTION_EXAMPLE2}, {ACTION_EXAMPLE3}).
3. Outcomes: For each {ACTION}, set the outcome type ("str" or "float") and define states that rule out certain {TRUTH}s. Avoid strict 1-to-1 mappings; states should rule out multiple {TRUTHS}. For example:
   - A {STATE_EXAMPLE1} on {ACTION_EXAMPLE1} might rule out {TRUTH_EXAMPLE1} and {TRUTH_EXAMPLE2}.
4. Ensure logical relationships between {TRUTH}, {ACTION}, and outcomes.

Example Format:
```python
Truths = [# List of {TRUTHS}]
Actions = [# List of {ACTIONS}]
Outcomes = {"Test Name": {"type": "str or float", "states": {"Outcome State 1": set(),  # Ruled-out {TRUTH} "Outcome State 2": set(), ...}},...}

Ensure:
- Tuples represent float types (e.g., (85, 100)); do not use inf.
- Allow empty sets.
- Include both float and string outcomes.
- Each test has at least 2 outcome states.
- Generate at least 30 actions and 50 truths, and each action must appear in Outcomes.
- The set of truths for each state MUST be ruled out by that outcome.
\end{lstlisting}
\end{tcolorbox}
\caption{Seed Config Generation Template}
\label{supp:fig2}
\end{figure}

\newpage

\begin{figure}[H]
\centering
\begin{tcolorbox}[
title=\textbf{Exemplar Config Generation Prompt},
colback=SeaGreen!10!CornflowerBlue!10,
colframe=RoyalPurple!55!Aquamarine!100!,
breakable,
fontupper=\sffamily
]
\begin{lstlisting}
Generate a configuration in Python for a Transdimensional Entity Identification game. The goal is to determine the being disrupting the fabric of space-time based on test outcomes. Follow the format below.

Requirements:
1.	Truths: List traits of transdimensional entities for identification (e.g., “Dimensional Instability”, “Temporal Anomalies”, “Spatial Distortion”).
2.	Actions: List diagnostic tests (e.g., “Interaction Experiments”, “Dimensional Stability Monitoring”, “Entity Behavior Analysis”).
3.	Outcomes: For each test, specify the outcome type (“str” or “float”) and define states that rule out certain traits (avoid 1-to-1 mappings). For example: An “Unstable Reading” on “Dimensional Stability Monitoring” might rule out “Dimensional Instability” and “Spatial Distortion”.
4.	Maintain logical relationships between traits, tests, and outcomes.

Example Format:

Truths = [] # List of traits
Actions = [] # List of tests
Outcomes = {"Test Name": {"type": "str or float", "states": {"Outcome State 1": set(), "Outcome State 2": set(),...}},...}

Ensure:
- Use tuples for float types (e.g., (85, 100)); no inf.
- Allow empty sets.
- Both float and string types must exist.
- Each test has at least 2 outcome states.
- Each outcome should rule out multiple traits.
- Generate at least 30 actions and 50 truths; every action must appear in Outcomes.
- The set of truths for a state MUST be ruled out by that outcome.
\end{lstlisting}
\end{tcolorbox}
\caption{Exemplar Config Generation Prompt. The prompt is generated based on the domain proposal in Supplementary Data Fig.\ref{supp:fig1} and \ref{supp:fig2}.}
\end{figure}

\begin{figure}[H]
\centering
\begin{tcolorbox}[
title=\textbf{Knowledge book generation prompt template},
colback=SeaGreen!10!CornflowerBlue!10,
colframe=RoyalPurple!55!Aquamarine!100!,
fontupper=\sffamily
]
\begin{lstlisting}
Please write a chemical analysis guidebook that introduces the following chemical substances and experiments in natural language according to the following information.

Chemical substances: {truths}
Experiments: {actions}
Outcomes: {ta_mapping}

Requirements:
1. The sets defined under the state of “Outcomes” represent the chemical substances that must be **excluded or ruled out** when the corresponding state is observed.
For example,
    If the state of "Experiment1" is:
    "Experiment1":
    {
        "states": {
            "Outcome_1": ["Substances1", "Substances2"],
            "Outcome_2": ["Substances3", "Substances4"]
        }
    }  
    This means:
    - When “Experiment1” is performed and “Outcome_1” is observed, “Substances1” and Substances2 are ruled out (i.e., they are eliminated as possibilities).
	- This exclusion approach is applied instead of confirming or indicating which substances are valid or related.
    
2. Explain the chemical substances and tests in a clear, straightforward manner to ensure the context and relationships are easy to understand.

3. Ensure that all relevant information is fully communicated without omissions. Every "rule-out" rules should be presented clearly and cohesively.
\end{lstlisting}
\end{tcolorbox}
\caption{Knowledge Book Generation Prompt Template}
\end{figure}

\begin{figure}[H]
\centering
\begin{tcolorbox}[
title=\textbf{Knowledge book revision prompt template},
colback=SeaGreen!10!CornflowerBlue!10,
colframe=RoyalPurple!55!Aquamarine!100!,
fontupper=\sffamily,
]
\begin{lstlisting}
Please evaluate an existing knowledge book generated by LLM based on the following information:
{input section(specific truths, actions, outcomes)}
{input_description(description for the input section)}
Existing Knowledge Book is the knowledge book to be evaluated.

## **Evaluation criteria**
Here are some common error cases, please evaluate if there are error cases appeared in the existing knowledge book.
### **Logical error: mistaking exclusion or rule out relationships for confirmation relationships**
The logical relationship in TA_mapping means exclusion rather than confirmation. Some knowledge books mistake exclusion for confirmation.

### Missing outcomes and observations in TA_mapping
Every action and its all valid outcomes in TA_mapping should be illustrated clearly in knowledge book. However, some knowledge book miss some elements in TA_mapping, resulting in the inaccuracy of the knowledge book. If the outcome cannot exclude any truths(empty set), it is acceptable to omit it.

### Generate exclusion relationships based on its own knowledge instead of strictly following the relationships in TA_mapping
Some knowledge books do not strictly follow the exclusion rules in TA_mapping and generate new exclusion rules based on their own understanding.

### Ambiguous description
The knowledge book should clearly illustrate the exclusion relationships rather than use ambiguous descriptions.

To sum up, the overall evaluation criteria are:
* **logical exclusion rather than confirmation**
* **completely illustrate all information in TA_mapping**
* **strictly follow TA_mapping without introducing any self-generate content**
* **clear and definite description about excluded truths**

Please evaluate the existing knowledge book based on the above criteria: {existing_knowledge_book}
Note again TA_mapping: {ta_mapping}. Please analyze **every** TA_mapping and its corresponding part in the existing knowledge book based on the evaluation criteria and explain whether it satisfies the criteria. Illustrate your reasons and if it is accurate, please output `<ANSWER>True</ANSWER>`, else output `<ANSWER>False</ANSWER>`. If the final answer is **False**, please revise the existing knowledge book and wrap the complete revised knowledge book in <BOOK> ${{revised knowledge book}} </BOOK>. **Note**: Please maintain the original language style, only correct the errors in the existing knowledge book when generating the revised knowledge book.
\end{lstlisting}
\end{tcolorbox}
\caption{Knowledge Book Revision Prompt Template}
\end{figure}

\begin{figure}[H]
\centering
\begin{tcolorbox}[
title=\textbf{Exemplar environment file generation prompt},
colback=SeaGreen!10!CornflowerBlue!10,
colframe=RoyalPurple!55!Aquamarine!100!,
fontupper=\sffamily,
]
\begin{lstlisting}
Please generate the environment file based on the following template:

```python
{example_env}
```

Here's the domain for the environment:
{domain[0]}

You only need to revise: knowledge_book_prompt, system_prompt, "xxxEnV" in registry and class name, and "from env.data.xxx_data import Truths, Actions, Outcomes" based on the given domain.
\end{lstlisting}
\end{tcolorbox}
\caption{Exemplar environment file generation prompt}
\end{figure}

\newpage



\section{Supplementary tables}
\FloatBarrier

\begin{table}[htbp]
\centering
\scriptsize               
\setlength{\tabcolsep}{4pt} 
\begin{adjustbox}{max width=\textwidth}  
\begin{tabular}{rl rl rl}
\toprule
Age & Major & Age & Major & Age & Major \\
\midrule
26 & Software Engineering & 26 & Education                     & 21 & Law \\
20 & Archaeology           & 20 & Computer Science and Technology & 28 & Computer Software and Theory \\
24 & Cultural Industry Management & 29 & Engineering            & 20 & Computer Science and Technology \\
24 & Medicine              & 24 & Pharmacy                     & 25 & Software Engineering \\
21 & Transportation        & 23 & Computer Science and Technology & 19 & Computer Technology and Science \\
22 & E\textendash commerce & 22 & Accounting                   & 20 & E\textendash commerce \\
21 & Software Technology   & 19 & Internet Applications        & 22 & Accounting \\
25 & Electronic Information Engineering & 22 & Accounting        & 26 & Accounting \\
24 & Business Administration & 28 & Intelligent Manufacturing  & 20 & Law \\
22 & English               & 22 & English Studies              & 23 & Civil Engineering \\
21 & Music                 & 23 & Robotics Engineering         & 24 & Mechatronics \\
22 & Big Data Technology   & 27 & Cloud Computing Applications & 28 & Food Science and Engineering \\
24 & Digital Media Technology & 22 & Electronic Information    & 20 & Medicine \\
20 & Mathematics and Applied Mathematics & 26 & Philosophy       & 23 & International Business \\
23 & International Economics and Trade & 26 & Network Engineering & 27 & Design \\
24 & Resources and Environment & 22 & Cloud Computing Technology Applications & 25 & Business English \\
22 & Mathematics           & 21 & Control and Management Science & 23 & Machine Tool Automation \\
20 & Chinese Language and Literature & 19 & Internet and New Media & 23 & Bioengineering \\
21 & Electronic Information Engineering & 22 & Environmental Engineering & 26 & Environmental Design \\
24 & Accounting            & 22 & Pharmacy                    & 20 & Electrical and Electronic Engineering \\
28 & Software Engineering  & 24 & Marketing                   & 23 & Hotel Management \\
22 & Financial Engineering & 27 & Exhibition Economy and Management & 25 & Cloud Computing Applications \\
23 & Digital Media         & 21 & Digital Media Technology     & 20 & Software Technology \\
20 & Internet of Things Engineering & 24 & Engineering Management & 25 & Philosophy \\
26 & Accounting            & 19 & International Tourism       & 22 & Automatic Control \\
28 & Computer Network Technology & 24 & Marxism Studies         & 22 & International Economics and Trade \\
20 & Accounting            & 21 & Control and Management Science & 24 & Design \\
25 & Accounting            & 22 & Cloud Computing Technology Applications & 23 & Food Science and Engineering \\
21 & Materials Science and Engineering & 20 & Machine Tool Automation & 22 & Accounting \\
26 & Software Engineering  & 24 & Transportation              & 22 & Automatic Control \\
22 & Computer Science and Technology & 23 & Electrical and Electronic Engineering & 22 & Machine Tool Automation \\
20 & English Studies       & 23 & Education                   & 19 & Internet and New Media \\
19 & Internet of Things Engineering & 24 & Philosophy           &    &       \\[-0.5ex]
\bottomrule
\end{tabular}
\end{adjustbox}
\caption{Statistics of 92 human‐study participants}
\label{tab:age_major_compact}
\end{table}

\section{Supplementary experiments}
\subsection{Effectiveness of knowledge book revision}
To assess the necessity and impact of evaluating and revising knowledge books, we conduct a comprehensive set of experiments using both the original and the revised knowledge books. The evaluation is performed under two experimental settings: an easy setting involving 100 domains, and a hard setting involving 5 domains. The success rate achieved by DeepSeek-V3 is adopted as the representative evaluation result.

\begin{figure}[htbp]
    \centering
    \includegraphics[width=0.8\linewidth]{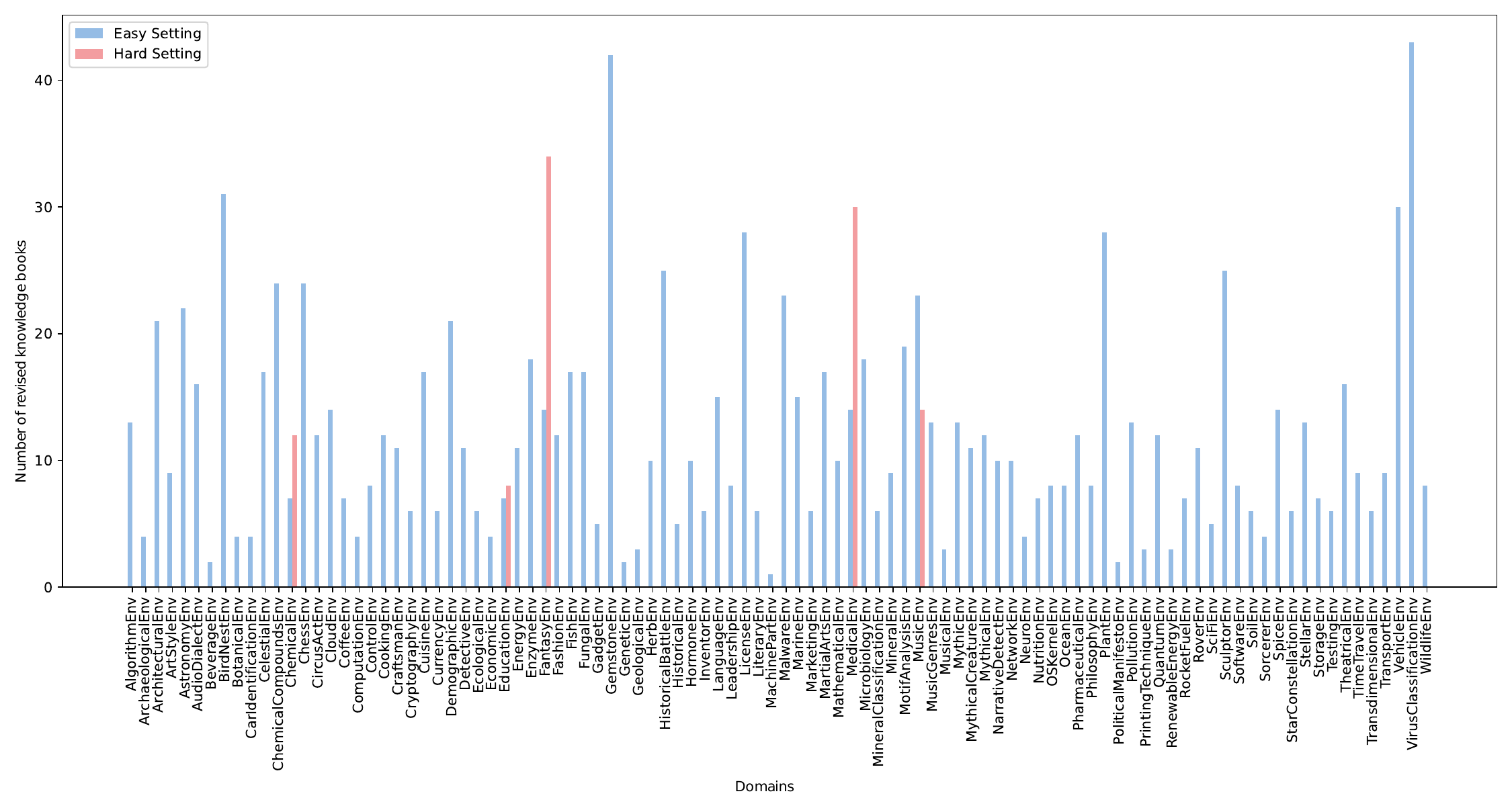}
    \caption{Number of revised knowledge books per domain in both easy and hard settings. Each domain initially contains 50 knowledge books in both settings. The horizontal axis denotes different domains. The vertical axis shows the number of revised knowledge books.}
    \label{fig:revise_num}
\end{figure}

First, after evaluating using the original knowledge books, we evaluate each knowledge book across all domains under both settings, and revise those that contain inaccuracies. We then calculate the number of revised books in each domain. As shown in Fig.\ref{fig:revise_num}, the number of revised knowledge books varies significantly across domains.

\begin{figure}[htbp]
    \centering
    \includegraphics[width=0.8\linewidth]{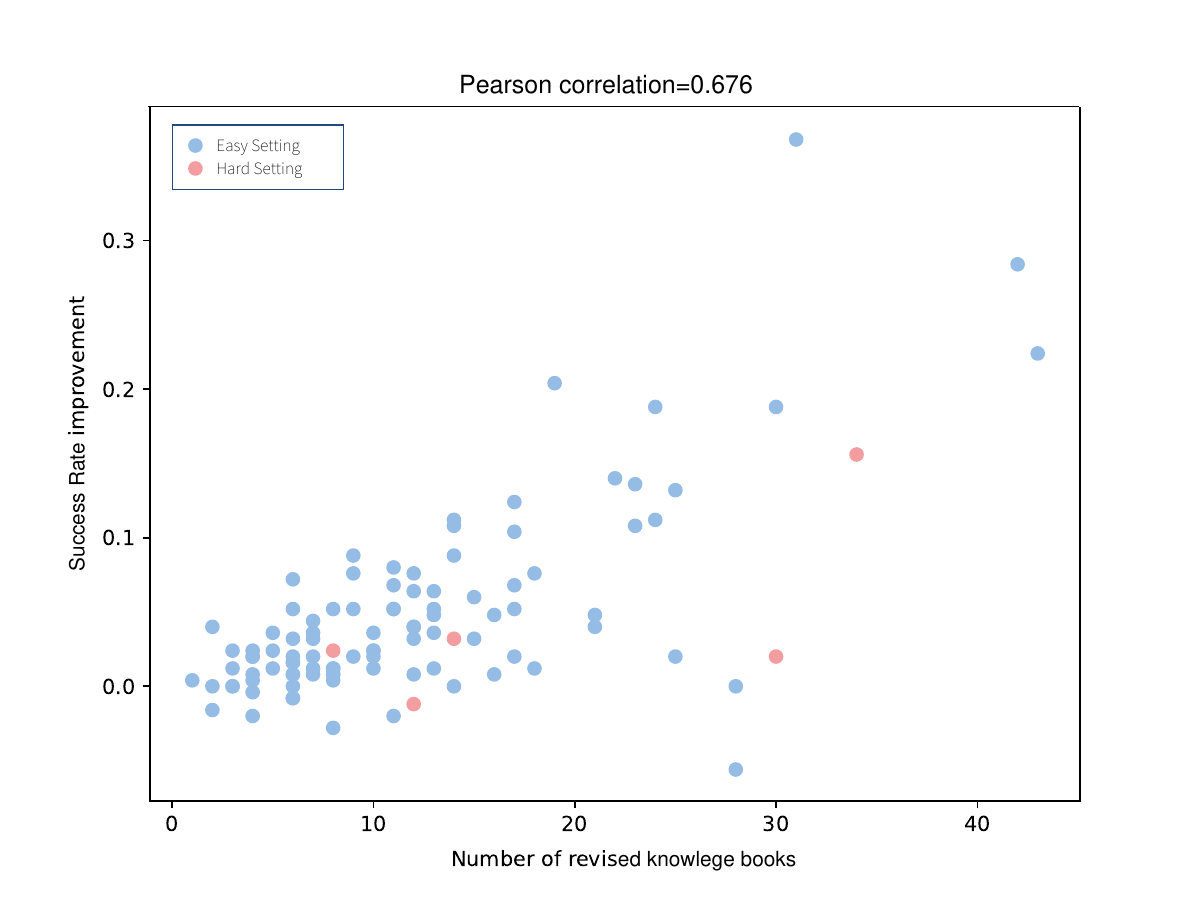}
    \caption{Correlation between the number of revised knowledge books and the corresponding improvement in success rate. Each point represents a domain, with results based on evaluations from DeepSeek-V3.}
    \label{fig:revise_correlation}
\end{figure}

Then we evaluate using the revised knowledge books and compare the success rate achieved using original knowledge books and revised knowledge books in each domain. The results are presented in Fig.\ref{fig:revise_correlation}.  A clear positive correlation can be observed between the number of revised knowledge books and the improvement in success rate. Specifically, the Pearson correlation coefficient is 0.676, indicating a strong positive relationship. These findings validate the effectiveness and necessity of knowledge book revision in achieving more accurate results by mitigating errors caused by inaccuracies in the knowledge books, rather than limitations in the LLM’s reasoning.

\subsection{Ablation study on configs and knowledge books generated by the DeepSeek series}
To veify the non-existence of potential bias stemming from employing a specific series of LLMs to generate configurations and knowledge books - since LLMs from the same series might potentially benefit from consistent internal knowledge and familiar linguistic patterns, we conduct an ablation study using a different LLMs series to generate configurations and knowledge books. In our primary experiment, we generate seed configs with o1, construct knowledge books using GPT-4o and subsequently revise knowledge book using o1-mini. Similarly, for the ablation study, we employ DeepSeek-R1 to generate seed configurations and revise knoledge books and DeepSeek-V3 for constructing knowledge books, given that both the configuration generation and book revision tasks entail substantial reasoning capabilities. We generate a representative environment: MedicalEnv and evaluate the representative LLM DeepSeek-V3 for the ablation configs generation series and GPT-4o for the previous configs generation series.

The experiment results are presented in Table.\ref{tab:ablation-study}. Although DeepSeek-V3 slightly outperforms GPT-4o in Success Rate, it simultaneously requires more actions to complete tasks. In GPT-generated environment, GPT-4o offers no clear success rate improvement over DeepSeek-V3. Both experimental settings reveal a consistent tend: DeepSeek-V3 is more accurate at identifying the correct ground truth, whereas GPT-4o tends to discover more efficient reasoning paths.  These mixed and modest differences all fall within expected experimental variability. We can conclude that there is no clear advantage for the LLM series generating the configs and knowledge books.
\begin{table}[ht]
\centering
\caption{Ablation Study Results in {MedicalEnv}.}
\begin{tabular}{llcccc}
\toprule
\multirow{2}{*}{{Environment}} 
  & \multirow{2}{*}{{Setting}} 
  & \multicolumn{2}{c}{{Success Rate}$\uparrow$} 
  & \multicolumn{2}{c}{{Relative Action Count}$\downarrow$} \\
\cmidrule(lr){3-4}\cmidrule(lr){5-6}
  &  & {DeepSeek-V3} & {GPT-4o} & {DeepSeek-V3} & {GPT-4o} \\
\midrule
\multirow{2}{*}{DeepSeek-generated} 
  & Easy & 0.884 & 0.812 & 0.636 & 0.571 \\
  & Hard & 0.584 & 0.468 & 1.722 & 1.592 \\
\midrule
\multirow{2}{*}{GPT-generated} 
  & Easy & 0.904 & 0.904 & 0.425 & 0.357 \\
  & Hard & 0.596 & 0.564 & 1.617 & 1.603 \\
\bottomrule
\end{tabular}
\label{tab:ablation-study}
\end{table}

\section{Pseudo code}

\begin{algorithm}[H]
\LinesNumbered
\SetAlgoLined
\SetNlSty{textbf}{}{}
\SetAlgoNoEnd
\caption{\textsc{SATSolver} Subroutine}
\SetKwInOut{Input}{Input}
\SetKwInOut{Output}{Output}

\Input{%
  $\mathcal{D}^{\text{valid}}$: set of outcomes (each with its list of excluded truths) that do not contradict $T^{\text{valid}}$,\\
  $T^{\text{invalid}}$: set of truths to be invalidated,\\
  $N^{\text{action}}$: maximum number of actions allowed
}
\Output{%
  A subset of actions $\text{actions}$ and a chosen outcome for each selected action (if any)
}

\BlankLine

\Commentm{\textbf{Create boolean variables}}
For each outcome $o_a$ in $\mathcal{D}^{\text{valid}}$, define a boolean variable $x_{a,o_a}$ indicating selection of $o_a$.

\Commentm{\textbf{Enforce constraints}}
\begin{enumerate}
  \item \emph{Unique state per action constraint:} For each action $a$, at most one $o_a$ can be selected. Formally, 
  \[
    \sum_{o_a \in \mathcal{D}^{\text{valid}}: \text{belongs to action } a} x_{a,o_a} \leq 1.
  \]
  \item \emph{Action limit constraint:} The total number of selected actions must not exceed $N^{\text{action}}$. If $A_{\text{chosen}}$ is the set of actions with at least one outcome selected, 
  \[
    |A_{\text{chosen}}| \leq N^{\text{action}}.
  \]
  \item \emph{Invalid truth exclusion constraint:} For every $t \in T^{\text{invalid}}$, there must be at least one selected outcome $o_a$ that excludes $t$. In terms of boolean variables,
  \[
    \forall \, t \in T^{\text{invalid}},\; \bigvee_{\substack{\\ o_a\;\text{excludes}\; t}} x_{a,o_a} = 1.
  \]
\end{enumerate}

\Commentm{\textbf{Solve the SAT formula}}
Use any standard SAT solver to solve for $\{x_{a,o_a}\}$. If a satisfying assignment is found, record the selected outcomes (those $x_{a,o_a} = 1$). The resulting \emph{actions} are those $a$ for which at least one $o_a$ is selected. Return $(\text{actions}, \text{outcomes})$. If unsatisfiable, indicate failure.

\end{algorithm}

\begin{algorithm}[t]
\LinesNumbered
\SetAlgoLined
\SetNlSty{textbf}{}{}
\SetAlgoNoEnd
\caption{SAT-based Task Generation}
\SetKwInOut{Input}{Input}
\SetKwInOut{Output}{Output}

\Input{%
  Universal truth set $T^{\text{univ}}$, 
  universal action set $A^{\text{univ}}$, 
  outcomes mapping $\mathcal{O}$, 
  symbolic knowledge book $K^{\text{symb}}$, 
  number of truths $N^{\text{truth}}$, 
  number of actions $N^{\text{action}}$, 
  number of valid truths $N^{\text{valid}}$, 
  required number of task instances $D$%
}
\Output{A collection of $D$ generated task instances}

\For{$i \leftarrow 1$ \KwTo $D$}{
  \Commentm{\textbf{Randomly select truths and designate valid/invalid truths}}
  $T^{\text{sub}} \leftarrow$ random subset of $T^{\text{univ}}$ of size $N^{\text{truth}}$\;
  $T^{\text{valid}} \leftarrow$ random subset of $T^{\text{sub}}$ of size $N^{\text{valid}}$\;
  $T^{\text{invalid}} \leftarrow T^{\text{sub}} \setminus T^{\text{valid}}$\;
  Initialize $\mathcal{D}^{\text{valid}} \leftarrow \emptyset$; \quad $A^{\text{rel}} \leftarrow \emptyset$\;

  \Commentm{\textbf{Assess outcomes for contradiction and relevance}}
  \ForEach{action $a \in A^{\text{univ}}$}{
    \ForEach{outcome $o_a \in \mathcal{O}_a$}{
    \Commentm{Obtain the truths excluded by $o_a$ from $K^{\text{symb}}$}
    \If{\emph{$o_a$ excludes any truth in $T^{\text{sub}}$}}{
    add $a$ to $A^{\text{rel}}$ \tcp*[r]{At least one of its outcomes excludes a truth in $T^{\text{sub}}$}
        \If{\emph{$o_a$ excludes any truth in $T^{\text{valid}}$}}{
        mark $o_a$ as \emph{contradictory} (cannot coexist with $T^{\text{valid}}$)\;
      }
      \Else{
        record $o_a$ and its excluded truths in $\mathcal{D}^{\text{valid}}$\;
      }
    }
    }
  }

  \Commentm{\textbf{Invoke SAT solver to select actions/outcomes}}
  $(\text{actions}, \text{outcomes}) \leftarrow \textsc{SATSolver}\bigl(\mathcal{D}^{\text{valid}}, T^{\text{invalid}}, N^{\text{action}}\bigr)$\;

  \If{\emph{the SAT solver reports unsatisfiable}}{
    \text{resample the $i^{th}$ task again} \tcp*[l]{Attempt again}
  }

  \Commentm{\textbf{Ensure we have $N^{\text{action}}$ actions}}
  \If{\emph{\#selected actions $<$ $N^{\text{action}}$}}{
    select additional actions from $A^{\text{rel}}$ (only outcomes not marked \emph{contradictory}), then from irrelevant actions if needed, until \#actions $= N^{\text{action}}$\;
  }

  \Commentm{\textbf{Check against duplication and save if unique}}
  \If{\emph{generated instance is not a duplicate}}{
    save the generated instance\;
  }
}
\end{algorithm}

\begin{algorithm}
\LinesNumbered 
\SetAlgoLined 
\SetNlSty{textbf}{}{} 
\SetAlgoNoEnd 
\caption{OptimalSearch(T, A)}
\label{algo_optimal_search}
\SetKwInOut{Input}{Input}
\SetKwInOut{Output}{Output}
\Input{Current Truth Space $T_{\text{current}}$, Current Action Space $A_{\text{cureent}}$, Truth-Action Mapping $M_{TA}$}
\Output{Minimum Expected Steps $E[S]$, Best Action $a^*$}

    \tcc{Check base cases: terminate the recursion if one of the base cases is satisfied }
    \If{$|T_{\mathrm{current}}| \leq 1$ \textbf{or} $|A_{\mathrm{current}}| = 0$}{
        \Return $0,\ \text{None}$\;
    }
    \For{each $t \in T_{\mathrm{current}}$}{
        \If{no $a \in A_{\text{current}}$ is related to $t$}{
            \Return $0,\ \text{None}$\;
        }
    }
    \tcc{Encode the current state into a bitmask $B$}
    $B \leftarrow \sum_{t \in T_{\text{current}}} 2^{\text{idx}(t)} + \sum_{a \in A_{\text{current}}} 2^{\text{idx}(a)}$\;
    \If{$B$ exists in $\mathrm{BestActionDict}$}{
        \Return BestActionDict[$B$]\;
    }
    Initialize $E[S]_{\text{min}} \leftarrow +\infty$, \quad $a^* \leftarrow \text{None}$\;
    \For{each $a \in A_{\mathrm{current}}$}{
        $A_{\text{next}} \leftarrow A_{\text{current}} \setminus \{a\}$\;
        \For{each state $s \in S_a$}{
            $W(s) \leftarrow |T_{\text{current}}| - |T_s \cap T_{\text{current}}|$\;
        }
        $Z \leftarrow \sum_{s \in S_a} W(s) + \epsilon$\;
        \For{each state $s \in S_a$}{
            $P(s) \leftarrow \dfrac{W(s)}{Z}$\;
        }
        $E[S]_a \leftarrow 0$\;
        \For{each state $s \in S_a$}{
            $T_{\text{next}} \leftarrow T_{\text{current}} \setminus T_s$\;
            $E[S]_{\text{next}},\ \_ \leftarrow \text{OptimalSearch}(T_{\text{next}}, A_{\text{next}})$\;
            $E[S]_a \leftarrow E[S]_a + P(s) \cdot E[S]_{\text{next}}$\;
            \If{$E[S]_a \geq E[S]_{\text{min}}$}{
                \textbf{break}\;
            }
        }
        \If{$E[S]_a < E[S]_{\text{min}}$}{
            $E[S]_{\text{min}} \leftarrow E[S]_a$, \quad $a^* \leftarrow a$\;
        }
    }
    BestActionDict[$B$] $\leftarrow (1 + E[S]_{\text{min}},\ a^*)$\;
    \Return $1 + E[S]_{\text{min}},\ a^*$\;

\end{algorithm}


\end{document}